%% file: main.tex
\def\AuthorVersion{1}
\documentclass[10pt]{article}
\ifdefined\AuthorVersion
  \usepackage[preprint]{tmlr} % Named manuscript; no publication/acceptance claim.
\else
  \usepackage{tmlr} % Anonymous TMLR review submission.
\fi
\newcommand{\PaperPDFTitle}{Learning Programs as Neural Networks for Generalization and Reuse}
\input{common}
\newtheorem{definition}{Definition}
\newtheorem{remark}{Remark}
\newtheorem{lemma}{Lemma}
\title{Learning Programs as Neural Networks for Generalization and Reuse}
\ifdefined\AuthorVersion
  \input{authors_en}
\else
  \author{Anonymous Authors}
\fi
\date{}

\begin{document}
\maketitle
\begin{abstract}
We investigate knowledge acquisition through human-like logical reasoning by learning executable computations from examples and retaining them for later use. Generalization is sought through acquiring a program that expresses the rule behind the examples. Neural NP4G represents programs as networks of computational nodes and directed connections, jointly learns their structure and numerical parameters, and accumulates acquired programs as reusable automatically defined functions (ADFs). We assume access to a detector that checks candidate behavior and guides acceptance or correction. We evaluate this approach on relational reasoning tasks with known specifications, where faithful logical extraction and equivalence checking prove that each accepted circuit implements its target on every input in the declared domain under real arithmetic. In a 160-run detector comparison, feeding the exact detectors' counterexamples back into training increased verified acquisition on both edge reversal and edge symmetrization. In the main experiment, selecting the first program verified by the equivalence oracle among ten seeds per task yielded one library covering all fourteen tasks. With the selected dependencies held fixed for downstream tasks, acquisition succeeded in 74/100 primary and 22/40 auxiliary attempts. The learned circuit for reachability within six edges composed two previously acquired relations, for paths of at most two and at most four edges, in the opposite order from the target formula. We proved that this circuit computes the target reachability relation on every finite graph in the declared domain.
\end{abstract}

\section{Introduction}
Our long-term goal is knowledge acquisition through human-like logical reasoning.\footnote{The authors used ChatGPT (OpenAI) and Claude (Anthropic) as assistive tools for research brainstorming, experimental code generation, theoretical analysis, and manuscript preparation, including translation and language editing. The authors independently reviewed and validated the research methodology, implementations, theoretical claims, and experimental results, and take full responsibility for the content of this work.} A learner should infer an executable computation from examples, apply it to unobserved inputs, and retain it as knowledge for subsequent tasks. We approach generalization through program acquisition: learning the operations and dependencies that generate the outputs. This makes acquired knowledge executable, available for inspection, and composable with earlier knowledge.

A program is represented as a directed network. Nodes perform computations, and connections specify which results become inputs to other computations. Selecting these components and connections determines the program's organization. An acquired network can then become a named component of another network. Learning consequently operates within a task, by constructing a computation, and across tasks, by expanding the repertoire of reusable programs. This is the program-acquisition concept of NP4G \citep{hara2024np4g}.

Neural NP4G realizes that concept with numerical neural units, weighted connections, and binary selection gates. Connections determine active computational paths; weights and biases determine the transformations along them. Joint optimization makes neural-network learning methods available for learning both program structure and numerical behavior. Acquired circuits are retained as automatically defined functions (ADFs), so later programs can build on earlier computations.

\paragraph{Detector hypothesis.} We assume that the learner has access to a mechanism that checks candidate behavior and returns acceptance, a discrepancy, or an inconclusive outcome. Its information and feedback are explicit parts of the learning setting. Angluin's minimally adequate teacher answers membership and equivalence queries and returns counterexamples \citep{angluin1987learning}; counterexample-guided inductive synthesis alternates candidate construction with verification and additional examples \citep{solarlezama2006sketching}. Cognitive error-monitoring research connects detected errors with compensatory behavior \citep{gehring1993error}, motivating the functional role of correction in this design.

We ask how to establish what a learned circuit computes on unseen structures and how that meaning carries through composition. Family relations and graphs provide targets involving existence, composition, bounded reachability, and counting. In the principal experiment, the learner uses examples to construct a reusable circuit. The detector holds the target formula and tests whether the formula faithfully extracted from the circuit is equivalent to that target. A second experiment varies the information available to the detector and whether its counterexamples return to training.

Selecting the first circuit proved equivalent to its target among ten seeds per task yielded one library covering all fourteen tasks. For each of the 37 circuits accepted on the six primary tasks requiring reuse, disabling any single mandatory ADF increased errors on independent checking data.

\section{Related Work}
\label{sec:related}
Network Programming for Generalization (NP4G) composes supplied functions and stored data in directed networks and reuses learned networks as ADFs \citep{hara2024np4g}. Its reported search randomly selects nodes and connections. Neural NP4G uses jointly optimized numerical units and gates. Topology and weight learning has precedents in NEAT, pruning, stochastic gates, and Deep Rewiring \citep{stanley2002neat,han2015connections,louizos2018lzero,bellec2018rewiring}; genetic programming and DreamCoder develop reusable program libraries \citep{koza1994gp2,ellis2021dreamcoder}.

Inductive logic programming (ILP) and Popper learn symbolic hypotheses from examples and background knowledge \citep{muggleton1991ilp,cropper2021popper}. Neural LP and differentiable ILP optimize relational rules \citep{yang2017neurallp,evans2018dilp}. Neural Logic Machines (NLM) learn predicate combinations with shared multilayer perceptrons. Their reported size generalization is measured on sampled structures larger than the training structures \citep[Appendix~B.6]{dong2019nlm}. Differentiable Logic Machines (DLM) extract discrete programs and reuse learned predicates \citep{zimmer2023dlm}. For each circuit Neural NP4G accepts, it checks that the saved numerical computation, including its reused dependencies, implements the specified rule throughout the declared domain.

Neural rule and automaton extraction provide explicit representations of learned computation \citep{towell1993rules,andrews1995survey,weiss2018automata}. Abstract interpretation supports our numerical enclosures \citep{cousot1977abstract}, and query containment supplies the logical comparison \citep{chandra1977cq,levy1993updates,wei2003negation}. GNN characterization and verification results motivate explicit architectural and logical restrictions \citep{barcelo2020logical,salzer2023limits,benedikt2024decidability}. Appendix~\ref{app:related} relates these methods to programmatic-policy extraction and contemporary systems combining learned proposals with executable or formal feedback.

\section{Problem Formulation}
\label{sec:setting}
\label{sec:verification-scope}
The relational evaluation uses finite nonempty structures with Boolean unary and binary predicates. A target $\varphi$ assigns truth labels to objects or object pairs; examples pair structures with these labels. The domain $\Gclass$ contains all structures satisfying declared predicate meanings and constraints, such as functional or disjoint relations. Object counts have no common upper bound. Appendix~\ref{sec:tasks} defines the experimental domains.

For a target of arity $r_{\mathrm{tar}}\in\{1,2\}$ and a structure $A$ with objects $U_A$, evaluated arguments are $D_1(A)=U_A$ and $D_2(A)=\{(x,y)\in U_A^2:x\ne y\}$. Binary ADF outputs are false on self-pairs. Let $\Circuit_\theta^{\mathbb R}$ denote the learned circuit under the exact-real interpretation of its saved binary parameters. The full-specification detector supplies the principal study's acceptance criterion:
\begin{definition}[Verified acquisition]
A circuit acquires its specified target when a formula $\psi_\theta$ is extracted and sound checks establish, for every $A\in\Gclass$ and $\bm x\in D_{r_{\mathrm{tar}}}(A)$,
\begin{align}
 \Circuit_\theta^{\mathbb R}(A,\bm x)&=\psi_\theta(A,\bm x),\label{eq:fidelity}\\
 \psi_\theta(A,\bm x)&=\varphi_A(\bm x).\label{eq:equivalence}
\end{align}
\end{definition}
The first equality is extraction faithfulness; the second is target equivalence. Acceptance returns the circuit and its logical description. The target guides continuation and selection, while gradients and local connection-removal decisions use training data. Mandatory reuse also requires the checks in Section~\ref{sec:reuse}.

\section{Neural NP4G}
\label{sec:method}
\label{sec:design}
Learning selects nodes, connections, and available operations while adjusting weights, biases, input embeddings, and readout parameters. The resulting circuit is an executable program; storing it as an ADF makes it available for later composition. This section defines its local computation and the relational execution used in the experiments. Circuit nodes are numerical units; the same computational graph is shared across objects in the input structure.

\subsection{Circuit structure and execution}
\label{sec:execution}
\label{sec:program}
\label{sec:ticks}
The candidate circuit has input, hidden, and output nodes $\mathcal I,\mathcal H,\mathcal O$. Every connection except one entering an input is eligible, including recurrent connections. Binary gates $g_{uv}$ and $g_v$ select edges and hidden nodes; input and output nodes remain active. With $\widehat w_{uv}=g_{uv}w_{uv}g_ug_v$ and $\sigma(z)=(1+e^{-z})^{-1}$, one synchronous \emph{circuit tick} computes
\begin{align}
 r_v^{\tau+1}&=b_v+\sum_u\widehat w_{uv}a_u^\tau,\label{eq:circuit-sum}\\
 a_v^{\tau+1}&=\begin{cases}
 x_v & v\in\mathcal I,\\
 g_v\sigma(r_v^{\tau+1}) & v\in\mathcal H,\\
 r_v^{\tau+1} & v\in\mathcal O.
 \end{cases}\label{eq:circuit-tick}
\end{align}
Each call clamps its inputs $x$, starts hidden and output values at zero, executes $K$ ticks, and returns $F_\theta(x)$. Fixing $K$ in advance lets us study circuit acquisition, verification, and reuse as ADFs at the same local depth across all fourteen tasks without also learning when a circuit should stop. At $K=1$ the local output is affine in $x$. A hidden node computes its sigmoid activation on the first tick and can pass it to the output on the second. We therefore choose $K=2$, the smallest depth at which a hidden nonlinearity can affect the output, and use 12 candidate hidden nodes. The local function then consists of one sigmoid hidden layer plus an affine input skip. Output-to-output connections can contribute to the affine term. Hidden-to-hidden and output-to-hidden connections can affect the local output from the third tick onward (Eq.~\eqref{eq:two-tick-reduction} in Appendix~\ref{app:execution-details}).

A \emph{message tick} is the outer relational update: gather information across objects, apply the shared $F_\theta$ at each object or pair, and threshold its outputs into the next state. Circuit values reset on each call; relational states persist across $T$ message ticks. Thus $\tau,K$ describe local execution, while $t,T$ describe relational propagation. Fixed execution budgets are part of the candidate program class. Object count changes the application and aggregation ranges, leaving the parameter count unchanged.

\subsection{Relational states and primitive operators}
\label{sec:relational}
\label{sec:operations}
Inputs $X^{(0)},X_i^{(1)},X_{ij}^{(2)}$ contain global, object, and pair features, including any ADF outputs. With learned affine maps and $h(z)=+1$ for $z\ge0$ and $-1$ otherwise, initialization is
\begin{align}
 S_i^0&=h\bigl(L_n(X_i^{(1)})+L_g(X^{(0)})\bigr),\\
 \mathsf Q_{ij}^0&=h\bigl(L_e(X_{ij}^{(2)})+L_s(X_i^{(1)})+L_t(X_j^{(1)})\bigr).
\end{align}
The principal model has four components per state; distinct source and destination maps preserve direction. Each message tick concatenates the current states and selected aggregation results, applies $F_\theta$, then applies $h$. A shared linear readout after $T$ ticks yields truth when its output is nonnegative (Figure~\ref{fig:relational-architecture}).

Primitive operators provide numerical evidence for existence, universal conditions, capped counts, and relational composition. With $\relu(z)=\max(0,z)$, define for a Boolean relation $R$ and state component $s$
\begin{align}
 q_{ij}&=\relu\bigl(R_{ij}\sigma(4s_j)-\tfrac12\bigr),\\
 E_i(R,s)&=\max_j q_{ij},\label{eq:exists}\\
 U_i(R,s)&=1-\max_j\relu\bigl(R_{ij}(1-\sigma(4s_j))-\tfrac12\bigr),\\
 N_i(R,s)&=\min\bigl(8,\textstyle\sum_jq_{ij}\bigr).\label{eq:count}
\end{align}
Unary updates use these operators in both relation directions and a global maximum. At hard states $s_j\in\{-1,+1\}$, contributions are either zero or $\sigma(4)-1/2>0$; false contributions cannot accumulate. The outputs are numerical evidence, with count cap $\kappa=8$, whose ranges extraction interprets logically.

All sums and maxima include every actual object, including $j=i$, and exclude padding. Relations determine their own aggregation ranges: adjacency selects neighbors, while the graph identity channel $I_{ij}=\mathbf1[i=j]$ selects the current object. Including $j=i$ therefore lets a circuit test its own state through the same gated interface. Identity's fixed degree of one also produces redundant candidate branches (Appendix~\ref{sec:failure-analysis}). Binary ADF self-pairs remain false before aggregation.

Binary updates concatenate pair states, their transposes, input relations, and selected compositions
\[
 J_{ij}^{cd}=\max_k\relu\bigl(\sigma(4\mathsf Q_{ik}^{c})\sigma(4\mathsf Q_{kj}^{d})-\tfrac12\bigr).
\]
The intermediate $k$ includes $i$ and $j$. Candidates include matching components $c=d$ and ordered pairs of the first two components. A single message tick can compose relations spanning several graph edges, including those supplied by ADFs.
\input{generated/relational_architecture.tex}

\subsection{Optimization}
For a gate parameter $z_e$, set $p_e=\sigma(z_e)$, $g_e=\mathbf1[p_e>1/2]$, and $\widetilde g_e=\sg(g_e-p_e)+p_e$. The stop-gradient operator $\sg$ leaves values unchanged and has zero derivative. Forward gates are binary; this straight-through estimator differentiates through $p_e$ \citep{bengio2013ste}. During the principal experiment's first 20,000 updates, states transition as $h_\alpha(z)=\alpha h(z)+(1-\alpha)\tanh z$, with $\alpha$ increasing to one and gradients through $\tanh$. Verification and evaluation use hard states.

Training minimizes class-balanced binary cross entropy plus penalties for dense connections, unnecessary candidates, ambiguous gates, and nonbinary states. During additional training, a loss on inputs sampled from extraction ranges penalizes small sign margins to help resolve interval signs. Pruning reduces the combinations of incoming signals that extraction must consider. When the current training minibatch is error-free just before pruning, a removal must keep it error-free and must not increase errors at any object count on a fixed training subset. If a circuit remains unaccepted after additional training but fits the entire training set, final compression tries removing individual hidden nodes, then tests up to 128 candidate edges. It retains removals that preserve that fit and resubmits the circuit for verification. Appendix~\ref{app:training} gives the losses, removal order, and numerical settings.

\subsection{Acquired programs and reuse}
\label{sec:reuse}
The \emph{primitive operators} $E,U,N,J$ transform current relational states at each message tick. An operator gate selects one such transformation. An ADF gate selects a Boolean input feature computed by a frozen program, and the local edge gates route signals within the circuit. ADFs execute once on the original structure in dependency order before state initialization. Binary outputs are false on self-pairs. Downstream gradients leave the ADF weights fixed. Their outputs remain input features throughout the message ticks and are never passed back as arguments to recursive ADF calls.

Registration stores the circuit, extracted formula, arity, input specification, dependencies, and checking records, linked by saved-circuit hashes. A later program selects and combines these computations; once accepted, it can itself become an ADF. For example, the six-edge reachability circuit receives the frozen two-edge and four-edge reachability relations and learns how to combine them with its own fixed capacity. Dependency execution contributes to its total cost.

The experiment prescribes candidate and mandatory ADFs. When they suffice to construct a target, they are computed first and direct downstream access to original relations is disabled; otherwise the required original inputs remain available. Each mandatory ADF must occur in the extracted formula, have an active gate, and increase errors when individually disabled on independent checking data. Its saved definition and verified meaning enter the downstream equivalence check. Appendix~\ref{sec:tasks} specifies dependencies and restrictions.

\section{Program Extraction and Verification}
\label{sec:verification}
The principal procedure returns the first candidate passing faithful extraction and target-equivalence checking; other outcomes invoke the prescribed continuation budget. Figure~\ref{fig:pipeline} in Appendix~\ref{app:training} summarizes the loop. Verification uses exact-real semantics for saved finite binary parameters.

\paragraph{Numerical extraction.} The extractor pairs a logical condition with a box of numerical intervals. Together the conditions cover every declared input, and each box encloses the circuit state whenever its condition holds. Outward rounding through arithmetic and nonlinear functions preserves enclosure at each tick. For capped aggregation, a proved positive lower bound $\alpha$ on nonzero contributions gives a count $k$ at which $k\alpha\ge\kappa$. Enumerating smaller counts and one saturated tail covers arbitrary finite object counts. The extractor succeeds only when the readout sign is determined for every case that has not been proved impossible. A nonnegative lower bound gives true and a negative upper bound gives false. The disjunction of the true cases is then faithful on all inputs (Lemma~\ref{lem:extraction}).

For example, the saved Has a father circuit from seed~1 has readout intervals of $[-8.4355,-8.4354]$ for an object with no father and $[15.3924,15.3925]$ for an object with a father. Their signs yield the rule $\exists y\,\mathrm{father}(x,y)$. Appendix~\ref{app:worked-example} traces these bounds from the saved weights. The library uses the earlier successful seed~0.

\paragraph{Logical comparison.} ADF substitution preserves argument order and binary self-pair guards, with fresh bound variables. The relational fragment admits unions of conjunctive queries with positive binary atoms, unary literals, equality, and inequality. In any counterexample to containment, choose a witness assignment for a satisfied source conjunction. Restrict the domain to the objects assigned to its free and existential variables, keep only the binary edges it requires, and preserve all unary truth values on those objects. The denials allowed in Appendix~\ref{sec:ucq} and the declared acyclicity constraints survive these restrictions. A witness to the target query in the smaller structure would also satisfy it in the original, since its binary atoms are positive. Enumerating all identifications of the source variables and complete unary assignments on the resulting objects therefore suffices. Testing containment in both directions establishes equivalence (Lemma~\ref{lem:containment}). For Boolean combinations of capped counts in one relation and direction, we instead compare truth tables through the largest cap, after which both tables are constant (Appendix~\ref{sec:ucq}).

\begin{remark}[Composition of the checks]
\label{sec:audit}
\label{prop:acquisition-correctness}
For fixed $K,T$, finite saved parameters, and the declared operations and logical fragments, every circuit accepted by the reference procedure satisfies $\Circuit_\theta^{\mathbb R}(A,\bm x)=\varphi_A(\bm x)$ for every $A\in\Gclass$ and every evaluated $\bm x$. Extraction proves circuit-to-formula faithfulness. Bidirectional containment for the relational fragment or equality of the saturated count tables proves formula-to-target equivalence. Substituting verified dependencies in order and simplifying with Boolean identities, declared input definitions, and containment-proved rewrites preserve the result.
\end{remark}
The implemented checks use outward bounds after each arithmetic operation and no threshold tolerance. Their three outcomes are proved equivalence, a confirmed domain-valid counterexample, and inconclusive. Appendix~\ref{app:verification} records implementation conformance tests.

\section{Experiments}
\label{sec:evaluation}
\subsection{Benchmarks and protocol}
The principal study uses NLM's five family and five graph task categories \citep{dong2019nlm}, plus four auxiliary tasks for reusable relations. Family inputs have typed, functional, acyclic parent relations; graphs are directed and loop-free, with vertex colors. Each of the fourteen tasks runs seeds 0--9. The first verified program in that order supplies the frozen downstream ADF, producing one selected library. Every required dependency is acquired within the study.

Training uses 256 structures at each of 10, 20, 30, and 40 objects. The principal circuit has $K=2$; three auxiliary tasks use $T=2$ and the others $T=1$. A check after 25,000 updates can trigger another 25,000 updates, a fixed checkpoint portfolio, and training-error-preserving compression. At most seven candidates are checked per attempt. Held-out predictions influence none of these decisions. All 140 attempts remain in the denominator; unsuccessful attempts retain one diagnostic candidate. Appendices~\ref{sec:tasks} and~\ref{sec:study} specify targets, dataset sizes, seed streams, and continuation rules.

\subsection{Acquisition and reuse}
\label{sec:main-results}
Verified acquisition succeeds in 74/100 primary and 22/40 auxiliary attempts, with descriptive 95\% Clopper--Pearson intervals $[64.3,82.3]\%$ and $[38.5,70.7]\%$ \citep{clopper1934limits}. Oracle-assisted selection from ten seeds yields a program for each of the fourteen tasks (Table~\ref{tab:acquisition-summary}). The Sister and Brother relations each have one successful seed. All ten downstream Has a sister attempts use the same selected Sister circuit, and all ten succeed.

\begin{table}[htbp]
\centering\small
\caption{Acquisition outcomes, ten seeds per task. Acquired, counterexample, and inconclusive partition each row. Zero-error counts selected checkpoints with no validation or test errors. Blocks are primary family, primary graph, and auxiliary tasks. Per-task 95\% intervals and complete diagnostics are in Appendix~\ref{sec:results}.}
\label{tab:acquisition-summary}
\begin{tabularx}{\linewidth}{@{}Yrrrr@{}}
\toprule
Task & Acquired & Counterexample & Inconclusive & Zero-error\\
\midrule
Has a father & $10/10$ & 0 & 0 & 10 \\
Has a sister & $10/10$ & 0 & 0 & 10 \\
Grandparent & $9/10$ & 0 & 1 & 9 \\
Uncle & $4/10$ & 2 & 4 & 4 \\
Maternal great-uncle & $2/10$ & 3 & 5 & 4 \\
\midrule
Adjacent to red & $10/10$ & 0 & 0 & 10 \\
Reachability within 4 edges & $10/10$ & 0 & 0 & 10 \\
Reachability within 6 edges & $2/10$ & 8 & 0 & 2 \\
Outdegree one & $9/10$ & 0 & 1 & 10 \\
Outdegree two & $8/10$ & 0 & 2 & 10 \\
\midrule
Sister relation & $1/10$ & 3 & 6 & 6 \\
Parent relation & $10/10$ & 0 & 0 & 10 \\
Brother relation & $1/10$ & 2 & 7 & 3 \\
Reachability within 2 edges & $10/10$ & 0 & 0 & 10 \\
\midrule
Primary total & $74/100$ & 13 & 13 & 79\\
Auxiliary total & $22/40$ & 5 & 13 & 29\\
\bottomrule
\end{tabularx}
\end{table}

All 37 acquired programs in the six primary tasks requiring ADFs passed the checks for every mandatory dependency. In each case the formula included that dependency, its gate was active, and disabling it alone increased errors on independent use-check data. The formula with the verified dependencies expanded was equivalent to the target. Seven programs were acquired after compression. Across 294 checkpoint checks, 34 intermediate checks reached a resource limit. Of the 44 unsuccessful attempts, 18 final diagnostic candidates had counterexamples. The other 26 completed numerical extraction and stopped at a logical obstruction. Negated binary atoms accounted for 14 cases, negated existential joins for nine, and unsupported counting syntax for three (Appendix~\ref{sec:results}).

\subsection{Prediction and acquired computations}
Of the 79 primary and 29 auxiliary attempts with zero validation and test errors, five primary and seven auxiliary attempts received inconclusive verdicts. Two of these twelve candidates, one for outdegree one and one for outdegree two, required a non-red vertex somewhere in the graph. All-red graphs constructed afterwards from the saved formulas exposed this error. An object with the specified outdegree satisfies the target even when every vertex is red. In a third candidate, substituting the fixed identity degree removed a redundant branch and yielded the correct target. Reach6 has 97.56\% pooled test-label accuracy and 2/10 verified acquisitions. At 500 objects all recorded Reach6 labels are positive, so constant true also scores 100\%. Appendix~\ref{sec:failure-analysis} gives the diagnostic cases, and Appendices~\ref{app:prediction-detail} and~\ref{sec:results} report class prevalence and structure-accuracy bounds.

The first accepted Reach6 circuit simplifies to
\[
 \operatorname{Reach}_4(x,y)\ \vee\ \exists z\,[\operatorname{Reach}_2(x,z)\wedge\operatorname{Reach}_4(z,y)],
\]
where $\operatorname{Reach}_k$ denotes one-to-$k$-edge directed reachability with false self-pairs. The target composes $\operatorname{Reach}_4\circ\operatorname{Reach}_2$; the learned program reverses that order. Bidirectional containment, after expanding dependencies with their self-pair guards, establishes equivalence.

Standard outputs apply justified simplification and provide named ADF definitions once, retaining raw formulas and rewrite provenance. The standard Reach6 body has 113 characters, or 488 with its reduced dependencies defined once each. Adjacent-to-red reduces from a 355,428-character raw extraction to a compact existential rule. Appendix~\ref{sec:formula-results} reports all fourteen selected outputs, expanded and shared-definition sizes, and the finite cross-check.

NLM reports graduation in 9/10 Uncle and 6/10 Reach6 attempts \citep[Table~4]{dong2019nlm}. Graduation records seeds meeting a training-loss criterion. Our 4/10 and 2/10 count circuits with proved target equivalence. NLM describes undirected spatial nearest-neighbor graphs. We sample edges on ordered pairs independently given each graph's edge probability. Appendix~\ref{sec:nlm} gives the training budgets and evaluation settings.

\input{generated/detector_experiment_en.tex}
\FloatBarrier

\section{Discussion and Limitations}
\label{sec:discussion}
\label{sec:limitations}
In the principal experiment the detector already holds the target rule. What the learner acquires from examples is a numerical implementation that can be reused in later programs. One application is to compose verified subqueries on relational data whose predicate meanings and integrity constraints match the declarations used in the proof. The exact black-box detector retains the target's semantic information even when its formula is hidden. Partial properties and finite tests supply less information and can accept incorrect rules. The detector experiment measures the effect on two local tasks with ten paired seeds and a small architecture.

The principal curriculum supplies operations, arities, execution budgets, and dependencies. Its downstream acquisition rates are conditional on one selected library. Estimating the probability and cost of building a complete library requires independent runs from an empty library, accounting for training, verification, failed attempts, and dependency execution. Class imbalance and only four 500-object structures per seed limit predictive extrapolation. Sparse graphs and boundary cases covering both classes would provide a more informative test.

At $K=2$, the local circuit reduces to one sigmoid hidden layer with an affine input skip (Eq.~\eqref{eq:two-tick-reduction}). Relational operators aggregate information across objects, and previously acquired circuits can be composed across tasks. Within one local call, however, a signal passing through two successive hidden nodes cannot reach the output. The principal study therefore leaves acquisition of deeper local computations untested. Measuring the effect of learning connections on acquisition rates requires a fixed-connectivity comparison with the same relational operators and hidden width. Comparisons with NLM and DLM require matched evaluation criteria. Ablations could quantify the contributions of gating, pruning, compression, identity, and reuse. Direct execution or compilation of a known formula is another practical comparison. Simplified outputs expose computations, while human comprehension and equal-task description lengths require dedicated evaluation.

The proof covers the saved parameters under exact real arithmetic. Extending it to PyTorch/CUDA execution requires bounds on rounding errors, particularly near decision thresholds. The extractor is sound and partial. The logical procedure is complete on its declared fragment when enumeration finishes. The proof relies on the checker implementation and its numerical libraries. Each extraction has a 6 GB memory limit and a 1,800-second time limit, so correct candidates can remain unproved.

Cognitive error monitoring motivates the detector's role in correcting a candidate. The experiments evaluate that role in a computational model trained with global gradients. Assessing its relation to human reasoning would require neural or behavioral measurements. A local audit matched the archived experimental records to their hashes. Independent data regeneration and end-to-end reproduction remain unverified (Appendix~\ref{app:evidence-scope}).
\FloatBarrier

\section{Conclusion}
Neural NP4G learned executable circuits from relational examples. Selecting the first verified circuit among ten seeds per task produced one library covering all fourteen tasks. With the selected dependencies held fixed for downstream tasks, acquisition succeeded in 74/100 primary and 22/40 auxiliary attempts. In the separate 160-run detector experiment, counterexample feedback raised exact-oracle acquisition from 5/10 to 8/10 for edge reversal and from 2/10 to 5/10 for symmetrization. The circuit learned for reachability within six edges composed two previously acquired relations, for paths of at most two and at most four edges, in the reverse of the target order. We expanded their definitions in the extracted formula and proved containment in both directions against the target. Under exact real arithmetic, the composed circuit thus computes the target reachability relation on every finite graph in the declared domain\label{sec:main-end}.

\bibliographystyle{tmlr}
\bibliography{references}

\clearpage
\appendix
\section{Training and Verification Settings}
\label{app:training}
The loss is
\begin{align}
 \mathcal L={}&\mathcal L_{\mathrm{BCE}}+0.01\mathcal L_{\mathrm{structure}}
 +0.01\mathcal L_{\mathrm{candidate}}+0.02\mathcal L_{\mathrm{binarize}}\notag\\
 &+0.05\mathcal L_{\mathrm{state}}+\lambda_{\mathrm{synthetic}}\mathcal L_{\mathrm{synthetic}}.
\end{align}
The cross-entropy term gives each class equal total weight when both labels occur in a minibatch. The structure term weights edge and hidden-node densities by $10^{-4}$ and $10^{-3}$. The candidate term penalizes gate uncertainty and active candidates, normalized by $\max(5,K_{\mathrm{cand}})$ for $K_{\mathrm{cand}}$ candidates. The binarization term makes structural and candidate gate choices less ambiguous. The state term averages penalties encouraging absolute state values of at least 0.99 and hidden-node pre-activations away from zero.

The synthetic term draws 512 inputs from the numerical ranges used for extraction and averages the largest $1/16$ of violations of sign margins and interval-width conditions. This is a CVaR-style objective emphasizing the largest losses \citep{rockafellar2000cvar}. Its weight is zero during the initial training stage and 0.3 during additional training. Using the numerical abstraction to shape learning is related to verification-aware training with interval bounds and abstract interpretation \citep{gowal2018ibp,mirman2018diffai}, here applied to sampled extraction ranges.
\begin{table}[htbp]
\centering\small
\caption{Shared numerical settings. Only the relational-update count uses the task-specific settings in Appendix~\ref{sec:tasks}.}
\begin{tabularx}{\linewidth}{@{}>{\raggedright\arraybackslash}p{61mm}Y@{}}
\toprule
Setting & Value\\
\midrule
State components / hidden nodes & 4 / 12\\
Circuit ticks per message tick & 2\\
Initializations & Ten seeds, 0--9\\
Optimizer / batch size & Adam / 8 structures\\
Gradient-norm limit & 1.0\\
Initial training stage & 25,000 updates, learning rate 0.005, target incoming degree 10\\
Additional training & 25,000 updates, learning rate 0.002, target incoming degree 3\\
State transition to binary values & Complete after 20,000 updates\\
Connection-removal interval & 500 updates initially; 1,000 during additional training\\
Verification candidates & Incoming degree at most 6; at most 4 circuits\\
Extraction limits & 6 GB, 1,800 seconds; at least 3 GB free at start\\
Circuit-versus-formula checks & Small structures with 6, 9, 13, 20 objects\\
Independent counterexample search & 6 structures each at 50, 100, 200, 500 objects\\
\bottomrule
\end{tabularx}
\end{table}
\FloatBarrier

During additional training, single-edge removals are tested directly on synthetic inputs. A removal must reduce the corresponding violation by at least 75\%, with at most four locations changed per event. When the minibatch was error-free before pruning, the algorithm checks the proposed removals on that batch and on 16 fixed training structures per object count. If the joint proposal fails the criteria in Section~\ref{sec:method}, it is rolled back and up to 128 edges are tested individually in increasing order of absolute effective weight.

Initial gate parameters are $-2$ for most edges, $+2$ for input-to-hidden and hidden-to-output connections, and zero for input-to-output and hidden-to-hidden connections. Hidden-node gate parameters start at $+2$. Random-number streams are separated by use. Records identify the implementation, settings, environment, data, and saved circuits for consistency checks. Distribution documentation provides execution commands and the mapping to recorded evidence files.

\FloatBarrier
\input{generated/method_details_en.tex}
\FloatBarrier

\section{Verification Implementation}
\label{app:verification}
Each addition and multiplication is a separate binary64 operation, followed by moving to the next representable value in the required outward direction. Under IEEE binary64 arithmetic with gradual underflow, these endpoints enclose the exact operation. Exact zero, identity, and cancellation cases retain their values. Affine sums apply this rule after every product and addition, without a BLAS reduction or a final-only correction. Sigmoid and tanh use MPFR's directed rounding with binary64 exponent and subnormal ranges. Sigmoid uses a nonpositive exponent argument and enclosing directions separately for exponentiation, the denominator, and division. Clustering preserves original endpoints and merges only by an enclosing hull. Threshold decisions use no tolerance; nonfinite values and unsupported operations cannot yield acceptance.

The logical implementation enumerates all source-variable identifications and unspecified unary assignments, retaining explicit true and false values. Constraint checks reject structures without changing them; returned counterexamples are checked against the source query. A sufficient shortcut searches for a homomorphism whose target literals follow from source literals, including justified equality and typing consequences. Failure invokes the exhaustive procedure. Local simplification accepts $e\equiv\bot$ only after proving $e\subseteq\bot$, and $e\equiv\top$ only after proving $\neg e\subseteq\bot$; unsupported negations cause no rewrite. The degree checker derives its final region from every atomic saturation point. Enumeration limits return an inconclusive outcome, preserving soundness but limiting completeness of the executable checker.

Regression checks cover exact rational affine extrema and cancellation, subnormal products, high-precision nonlinear values, clustering containment, and threshold-straddling cases. Logical checks compare small-query containment against exhaustive finite models, preserve explicit negative types, and detect disagreement at degree 33 in $\min(\deg,33)\in\{1,33\}$ versus $\deg=1$. These checks test implementation conformance; the all-input argument is the enclosure induction and containment proof in Appendix~\ref{app:extraction-checking}. 

Each dependency was learned within the complete experiment and verified before reuse. Saved-circuit hashes link the selected training state, its extracted formula, dependency identities, and prediction reports. The final evidence audit checked these links and all trial and prediction denominators without changing the saved circuits or their selection.

\FloatBarrier
\section{Prediction Results by Object Count and Seed}
\label{app:prediction-detail}
Table~\ref{tab:test-sizes} reports diagnostic test label accuracy separately at each object count, pooling the selected checkpoints from ten acquisition attempts within a task and size. Each size therefore has its own denominator from Table~\ref{tab:data}. Table~\ref{tab:test-seeds} summarizes the ten individual seed accuracies, each pooling that seed's test labels, and bounds the proportion of completely correct structures. Both tables retain checkpoints from attempts that returned no program and condition downstream attempts on the same selected library. 

\begin{table}[htbp]
\centering\footnotesize
\setlength{\tabcolsep}{3pt}
\caption{Test label accuracy (\%) at each object count, pooled over ten seeds. Auxiliary tasks are marked with an asterisk. Displayed values are rounded to two decimals; the accompanying evidence retains integer error counts.}
\label{tab:test-sizes}
\begin{tabularx}{\linewidth}{@{}Yrrrrrrrr@{}}
\toprule
Task & $10$ & $20$ & $30$ & $40$ & $50$ & $100$ & $200$ & $500$\\
\midrule
Has a father & 100.00 & 100.00 & 100.00 & 100.00 & 100.00 & 100.00 & 100.00 & 100.00 \\
Has a sister & 100.00 & 100.00 & 100.00 & 100.00 & 100.00 & 100.00 & 100.00 & 100.00 \\
Grandparent & 93.31 & 92.00 & 91.44 & 91.14 & 90.94 & 90.50 & 90.26 & 90.11 \\
Uncle & 74.06 & 72.67 & 71.73 & 71.45 & 70.99 & 70.42 & 70.18 & 70.01 \\
Maternal great-uncle & 90.15 & 87.73 & 87.16 & 87.03 & 87.14 & 87.65 & 88.38 & 89.21 \\
Adjacent to red & 100.00 & 100.00 & 100.00 & 100.00 & 100.00 & 100.00 & 100.00 & 100.00 \\
Reachability within 4 edges & 100.00 & 100.00 & 100.00 & 100.00 & 100.00 & 100.00 & 100.00 & 100.00 \\
Reachability within 6 edges & 81.28 & 88.35 & 90.94 & 94.29 & 94.49 & 96.64 & 98.14 & 100.00 \\
Outdegree one & 100.00 & 100.00 & 100.00 & 100.00 & 100.00 & 100.00 & 100.00 & 100.00 \\
Outdegree two & 100.00 & 100.00 & 100.00 & 100.00 & 100.00 & 100.00 & 100.00 & 100.00 \\
Sister relation$^{*}$ & 82.59 & 81.93 & 81.67 & 81.36 & 81.24 & 80.75 & 80.44 & 80.21 \\
Parent relation$^{*}$ & 100.00 & 100.00 & 100.00 & 100.00 & 100.00 & 100.00 & 100.00 & 100.00 \\
Brother relation$^{*}$ & 76.28 & 75.58 & 75.35 & 75.26 & 75.23 & 75.26 & 75.14 & 75.39 \\
Reachability within 2 edges$^{*}$ & 100.00 & 100.00 & 100.00 & 100.00 & 100.00 & 100.00 & 100.00 & 100.00 \\
\bottomrule
\end{tabularx}
\end{table}

For a single seed and object count, let $n_{\mathrm{str}}$ be the number of structures, $\ell_{\mathrm{lab}}$ the number of labels in each structure, and $e_{\mathrm{err}}$ the total erroneous labels. If $n_{\mathrm{exact}}$ structures are completely correct, then
\[
 \max(0,n_{\mathrm{str}}-e_{\mathrm{err}})\ \leq\ n_{\mathrm{exact}}\ \leq\ n_{\mathrm{str}}-\left\lceil e_{\mathrm{err}}/\ell_{\mathrm{lab}}\right\rceil.
\]
The lower bound distributes errors over as many structures as possible, and the upper bound concentrates them in as few as possible. We sum these bounds over the separate seed--size groups, whose structure membership is known. These are deterministic bounds from the saved error counts, not confidence intervals or measurements of where errors occurred. When $e_{\mathrm{err}}=0$, both bounds give $n_{\mathrm{exact}}=n_{\mathrm{str}}$. For six-edge reachability, the records identify only a range of 1,359--5,622 completely correct structures among 6,280 tested across seeds. Recovering the exact count requires new inference on the original structures, retaining each structure's error count.

\begin{table}[htbp]
\centering\small
\caption{Distribution of test label accuracy across ten seeds, and bounds on completely correct test structures (all values in \%). Every task contributes 6,280 structures. Accuracies are rounded to two decimals; bound endpoints are rounded outward. A single value in the last column means the bounds coincide; a range is not an observed exact rate.}
\label{tab:test-seeds}
\begin{tabularx}{\linewidth}{@{}Yrrr>{\raggedleft\arraybackslash}p{34mm}@{}}
\toprule
Task & Minimum & Median & Maximum & Completely correct structures\\
\midrule
Has a father & 100.00 & 100.00 & 100.00 & 100.00 \\
Has a sister & 100.00 & 100.00 & 100.00 & 100.00 \\
Grandparent & 4.45 & 100.00 & 100.00 & 90.00--91.69 \\
Uncle & 5.35 & 94.86 & 100.00 & 40.00--71.74 \\
Maternal great-uncle & 8.57 & 95.52 & 100.00 & 40.00--87.55 \\
Adjacent to red & 100.00 & 100.00 & 100.00 & 100.00 \\
Reachability within 4 edges & 100.00 & 100.00 & 100.00 & 100.00 \\
Reachability within 6 edges & 90.89 & 98.70 & 100.00 & 21.64--89.53 \\
Outdegree one & 100.00 & 100.00 & 100.00 & 100.00 \\
Outdegree two & 100.00 & 100.00 & 100.00 & 100.00 \\
Sister relation$^{*}$ & 1.45 & 100.00 & 100.00 & 60.00--81.45 \\
Parent relation$^{*}$ & 100.00 & 100.00 & 100.00 & 100.00 \\
Brother relation$^{*}$ & 2.86 & 97.14 & 100.00 & 30.00--75.10 \\
Reachability within 2 edges$^{*}$ & 100.00 & 100.00 & 100.00 & 100.00 \\
\bottomrule
\end{tabularx}
\end{table}
\FloatBarrier
\input{generated/worked_example_en}
\FloatBarrier
\input{notation_en}

\FloatBarrier
\section{Extraction and Logical Checking Details}
\label{app:extraction-checking}
\subsection{Interval-based formula extraction}
\label{sec:interval}
The verification target must first be fixed. Interpret finite saved weights and biases as the exact rationals represented by their binary encodings, and execute the operations of Section~\ref{sec:method} over the reals. Denote this circuit by $\Circuit_\theta^{\mathbb R}$. Gates are saved zeros or ones, and state thresholding returns $+1$ at $z\geq0$. This differs from the rounded PyTorch execution $\Circuit_\theta^{\mathrm{fp}}$. The reference procedure concerns the former. A guarantee for the latter additionally requires a correspondence argument covering operation order, precision, nonlinear functions, and accumulated summation error.

An interval $I=[l,u]$ denotes $\gamma(I)=\{z\in\mathbb R:l\leq z\leq u\}$. Following abstract interpretation \citep{cousot1977abstract}, an abstract state is a finite collection of pairs $(\chi,\mathcal B)$, interpreted at each relevant object or object pair. Here $\chi$ is a logical condition on the input structure and $\mathcal B=\prod_c I_c$ is a box of component intervals. We maintain (a) coverage: at least one $\chi$ holds on every input; and (b) enclosure: whenever $\chi$ holds, the actual state lies in $\mathcal B$. Shared conditions retain some correlations between components; intervals alone do not represent all such correlations.

Write $\operatorname{rd}$ and $\operatorname{ru}$ for enclosing bounds rounded downward and upward. Intermediate arithmetic must also round outward; the following endpoints bound the exact real expressions:
\begin{align}
 [l,u]+[a,b]&=[\operatorname{rd}(l+a),\operatorname{ru}(u+b)],\\
 [l,u]-[a,b]&=[\operatorname{rd}(l-b),\operatorname{ru}(u-a)],\\
 [l,u]\,[a,b]&=[\operatorname{rd}(\min S),\operatorname{ru}(\max S)],
 \quad S=\{la,lb,ua,ub\}.
\end{align}
For a positive interval, $[l,u]^{-1}=[\operatorname{rd}(1/u),\operatorname{ru}(1/l)]$. With $w_i^+=\max(w_i,0)$ and $w_i^-=\min(w_i,0)$, an affine map has enclosure
\begin{equation}
 \left[
 \operatorname{rd}\!\left(b+\sum_i(w_i^+l_i+w_i^-u_i)\right),
 \operatorname{ru}\!\left(b+\sum_i(w_i^+u_i+w_i^-l_i)\right)
 \right].
 \label{eq:validated-affine}
\end{equation}
Every product and sum must preserve enclosure. Rounding only the final result of an ordinary matrix multiplication does not implement this rule.

Monotone $\exp$, sigmoid, and ReLU are enclosed through endpoint evaluation. Sigmoid is evaluated as $\sigma(I)=(1+\exp(-I))^{-1}$, using interval arithmetic at every step. The enclosure of max is $[\max_i l_i,\max_i u_i]$, and min is analogous. State thresholding gives $[1,1]$ when $l\geq0$, $[-1,-1]$ when $u<0$, and otherwise remains undecided. NaNs, unsupported operations, or numerical steps without guaranteed enclosure cannot be accepted. MPFR's directed rounding provides a way to construct these endpoint enclosures \citep{mpfrmanual}.

For the contributions $q_{ij}$ of Section~\ref{sec:operations}, interval evaluation must establish exact zero when the condition fails and an enclosure $[\alpha,\beta]$, $0<\alpha\leq\beta$, when it holds. A threshold-straddling case remains undecided. Existential aggregation gives $[0,0]$ for zero satisfying objects and $[\alpha,\beta]$ for at least one. Universal aggregation applies the same reasoning to violations and subtracts their maximum from one. For relation composition, products with an inactive argument must have upper bounds at most the threshold, while active--active products must have lower bounds above it. The result then reduces to existential aggregation.

Evaluate each aggregation contribution under its input conditions $\chi_r(\bm x,z)$. Require every branch to give either $[0,0]$ or an interval with positive lower bound. Let $b(\bm x,z)$ be the disjunction of conditions of positive branches. Coverage and enclosure establish that $b$ holds exactly when the actual contribution is positive. Take $\alpha$ to be the smallest positive-branch lower bound and $\beta$ the largest upper bound. If there is no positive branch, the maximum and sum of contributions are identically zero. Existence and composition use the two conditions $\exists z\,b$ and its negation; universal aggregation applies the same construction to violations.

For counting, let $n_{+}(\bm x)=|\{z:b(\bm x,z)\}|$. The condition for exactly $n_{+}$ objects requires pairwise distinct $z_1,\ldots,z_{n_{+}}$ satisfying $b$, and requires every $z$ satisfying $b$ to equal one of them. For $n_{+}=0$, it is $\neg\exists z\,b$; for at least $k$, only $k$ distinct witnesses are required. Intermediate conditions may therefore contain negation and universal quantification. If the final formula cannot be transformed with preserved meaning into the checking grammar below, return undecided.

A sum capped at $\kappa=8$ has enclosure $[\min(\kappa,n_{+}\alpha),\min(\kappa,n_{+}\beta)]$ for $n_{+}$ positive contributions. Choose an integer $k$ for which $k\alpha\geq\kappa$ is established, and enumerate $n_{+}=0,\ldots,k-1$ and $n_{+}\geq k$. The latter region gives $[\kappa,\kappa]$. Thus real arithmetic admits finitely many cases without bounding the number of objects. These bounds cannot simply be reused for a floating-point implementation of $n_{+}$ additions.

The reference extractor proceeds as follows. Exhausting numerical precision or enumeration budgets returns an undecided result.
\begin{enumerate}[leftmargin=*]
 \item Fix saved binary gates and enumerate all Boolean combinations of the inputs to initialization. Each combination defines a logical condition $\chi$. Evaluate initialization by interval arithmetic. An undecided state retains an interval enclosing both possibilities; it is not assigned a single truth value.
 \item At each message tick, construct inputs from the current component intervals, original Boolean predicates, and the aggregation cases above. Outputs of ADFs are Boolean predicates; their correspondence to their circuits is checked separately in dependency order.
 \item Enumerate products of input cases and conjoin their conditions. Clamp input nodes, initialize other nodes to zero, and perform exactly $K$ synchronous circuit ticks using Eq.~\eqref{eq:validated-affine} and the activation enclosures.
 \item Pass output intervals to the next state. If intervals are merged, take the minimum and maximum of their original endpoints and disjoin their conditions. Nearest rounding that shrinks an interval is forbidden. Each original case interval must be wholly included in its assigned enclosure.
 \item Remove a combination only when its condition is proved identically false. Allowed identities include $\chi\wedge\neg\chi=\bot$, constant simplification, and empty intersections of allowed sets for the same counted quantity. Retain cases whose satisfiability is unresolved. Absence from sampled inputs does not justify removal.
 \item Repeat for the prescribed $T$ message ticks, then enclose the readout under each remaining condition. Classify as true if the lower bound is nonnegative, false if the upper bound is negative, and undecided otherwise. The extracted formula is the disjunction of the conditions classified true.
\end{enumerate}
Spurious combinations in a Cartesian product can increase undecided outcomes; retaining them does not cause false acceptance. An optimization that groups inputs must enclose their union. The procedure unrolls fixed $K,T$ and does not assume convergence to a fixed point.

\begin{lemma}[Enclosure and faithfulness of reference extraction]
\label{lem:extraction}
If reference extraction using the specified endpoint operations accepts, its formula agrees with $\Circuit_\theta^{\mathbb R}$ on every $A\in\Gclass$.
\end{lemma}
\begin{proof}
Enumeration covers all Boolean inputs to initialization. Endpoint arithmetic and monotonicity enclose each local operation. Aggregation bounds cover every finite number of positive contributions. Induction on circuit and message ticks therefore preserves properties (a) and (b). Hull merging enlarges represented sets; removing identically false conditions loses no reachable state. Each accepted final case has the same truth value as the circuit, and coverage establishes agreement on all inputs.
\end{proof}
This is a sound partial procedure, not a complete extractor: even a correct circuit can remain undecided because of lost correlations or coarse intervals.

\subsection{Logical equivalence checking}
\label{sec:ucq}
Expand the definitions of ADFs in dependency order and replace derived input predicates by their definitions. Use fresh bound variables at each substitution, preserving argument order and false self-pairs in the outputs of binary ADFs. Rewrites based on assumptions $\Sigma$ must be equivalences under those assumptions. For example, $R\cap S=\varnothing$ licenses $R(x,y)\wedge\neg S(x,y)\equiv R(x,y)$. An unsupported negative literal cannot simply be removed.

Starting from unions of conjunctive queries (UCQs), we admit unary negation, equality, and inequality only as specified below.
Over finite nonempty domains with finite unary and binary vocabularies, use
\begin{align}
 \mathcal Q(\bm x)&::=\bigvee_i\exists\bm y_i\,\bigwedge_j L_{ij},\\
 L&::=R(v,w)\mid U(v)\mid\neg U(v)\mid v=w\mid v\ne w
       \mid\top\mid\bot.
 \label{eq:ucq-grammar}
\end{align}
Here $v,w$ are variables, with one or two designated free arguments. Negation applies only to base unary predicates. Negated binary predicates, negated existential formulas, general universal formulas, and function symbols are excluded. A count lower bound of $k$ can enter this fragment when it is expanded into $k$ pairwise-distinct existential witnesses.

Let $\Sigma$ be a finite set of constraints $\forall\bm v\,(\eta\Rightarrow\bot)$, where $\eta$ contains positive binary atoms, positive or negative unary atoms, equalities, and inequalities. For example,
\[
 R(x,y)\wedge R(x,z)\wedge y\ne z\Rightarrow\bot,\qquad
 R(x,y)\wedge\neg U(y)\Rightarrow\bot
\]
specify functionality and typing. Disjointness and irreflexivity have the same form. We also allow an explicitly declared acyclicity condition on a binary relation or a named union of relations: compute its transitive closure and reject any diagonal entry. Both these denials and acyclicity are preserved by restricting the domain and deleting binary edges. This preservation is the property used in the proof below. Dependencies requiring the existence of new objects or edges are excluded.

Decide $\mathcal Q_1\subseteq_\Sigma \mathcal Q_2$ by the following finite enumeration.
\begin{enumerate}[leftmargin=*]
 \item For each conjunction $q$ in $\mathcal Q_1$, enumerate every partition of its variables, including free variables. Interpret each block as one object and discard partitions violating an equality or inequality of $q$.
 \item On each quotient domain, include precisely the binary edges required by $q$. Enumerate every truth assignment to every unary predicate on every object. Discard structures that violate the positive or negative unary atoms of $q$ or any constraint in $\Sigma$. Do not subsequently change a truth assignment.
 \item On each remaining structure, fix the designated free arguments and enumerate all assignments to the existential variables of $\mathcal Q_2$. If $\mathcal Q_2$ is false, return the structure as a counterexample to containment. If every structure satisfies $\mathcal Q_2$, accept containment.
\end{enumerate}
Equivalence applies this test in both directions. If binary evaluation excludes $x=y$, conjoin the inequality to both queries. For $n_{\mathrm v}$ variables in $q$ and $n_{\mathrm u}$ unary predicates, there are at most $\operatorname{Bell}(n_{\mathrm v})2^{n_{\mathrm u}n_{\mathrm v}}$ candidate structures, where $\operatorname{Bell}(n_{\mathrm v})$ counts partitions of $n_{\mathrm v}$ elements. This bound comes from the query, not from testing input structures only up to a chosen object count.

\begin{lemma}[Containment on the declared fragment]
\label{lem:containment}
When enumeration completes, the reference algorithm is sound and complete for containment of Eq.~\eqref{eq:ucq-grammar} under $\Sigma$.
\end{lemma}
\begin{proof}
A returned counterexample satisfies $\Sigma$ and $\mathcal Q_1$ but not $\mathcal Q_2$ by construction. Conversely, choose a satisfying conjunction $q$ and assignment in any counterexample to containment. Identify variables mapped to the same object, retain only their image and the binary edges required by $q$, and copy unary truth values exactly from the original structure. The resulting structure still satisfies $\Sigma$ and occurs in the enumeration. If it satisfied $\mathcal Q_2$, the same witness assignment would satisfy $\mathcal Q_2$ in the original structure: binary atoms are positive, and unary truth values and object identity are preserved. This contradicts the choice of counterexample. Every failure of containment is therefore discoverable. The argument extends ordinary conjunctive-query containment \citep{chandra1977cq} by explicitly retaining quotients and complete unary assignments.
\end{proof}
If additional input conditions cannot be expressed in $\Sigma$, containment over $\Sigma$ still holds on the more restricted input domain. A counterexample violating those additional conditions, however, does not establish a mismatch on that domain. Resource exhaustion is undecided, never acceptance.

For one relation and direction, let $\deg(x)$ be its degree. We allow Boolean combinations of atoms of the form
\[
 \min(\deg(x),k)\in\mathcal D,
\]
where $\mathcal D$ is a set of allowed counts. Let $k_{\max}$ be the largest atomic cap in either formula. Evaluate both formulas at degrees $0,\ldots,k_{\max}-1$ and in the region $\deg\geq k_{\max}$. Every atom is constant beyond $k_{\max}$, so equality of these tables is equivalent to equality at all nonnegative integer counts. A fixed cutoff requires checking that every atom saturates within it. General integer arithmetic is excluded.

\FloatBarrier
\section{Detailed Experimental Evidence}
\label{app:experiment-detail}
\subsection{Tasks and datasets}
\label{sec:tasks}
\label{sec:data}
The primary benchmark follows NLM's five family and five general-graph tasks \citep{dong2019nlm}. Family tasks infer relations from father, mother, son, and daughter inputs; graph tasks infer adjacency to a color, bounded reachability, and exact outdegree from edges and vertex colors. Table~\ref{tab:task-definitions} gives each target formula, required ADFs, and message tick count.

Four auxiliary tasks acquire ADFs for sister, parent, brother, and reachability within two edges. Some primary tasks also supply ADFs to later tasks. All 14 tasks are reported, but auxiliary tasks are excluded from the primary benchmark's denominator.

The experiments use two input domains. Family inputs are father, mother, son, and daughter. Each person has at most one father and one mother; fathers are male and mothers female. Sex categories are disjoint, and parenthood is acyclic. Son and daughter reverse the parent relation and condition on the child's sex. Graph inputs encode adjacency, identity, and vertex color, using directed graphs without self-edges. These declarations specify $\Gclass$ for the experiments.

The predicates $F(x,y)$ and $M(x,y)$ mean that $y$ is the father or mother of $x$, respectively; $P=F\vee M$ denotes parenthood. The predicates $\operatorname{son}(z,y)$ and $\operatorname{daughter}(z,y)$ mean that $y$ is a son or daughter of $z$. Sister $S$, brother $\operatorname{Bro}$, and uncle $V$ are defined by
\begin{align}
 S(x,y)&=(x\ne y)\wedge\exists z[M(x,z)\wedge\operatorname{daughter}(z,y)],\\
 \operatorname{Bro}(x,y)&=(x\ne y)\wedge\exists z[P(x,z)\wedge\operatorname{son}(z,y)],\\
 V(x,y)&=\exists z[P(x,z)\wedge \operatorname{Bro}(z,y)].
\end{align}
The sister task therefore concerns female siblings sharing a mother, while the brother task concerns male siblings sharing at least one parent. Both definitions use parent relations and sex categories, independently of relative age.

For graphs, let $\operatorname{adj}(x,y)$ denote an edge and $\operatorname{red}(y)$ the red property. The predicate $\operatorname{Reach}_k(x,y)$ means that distinct vertices $x,y$ are connected by a directed path of length between one and $k$. Binary formulas are evaluated on $D_2(A)$ from Section~\ref{sec:verification-scope}; self-pair outputs of ADFs are false.
\begin{table}[!htb]
\centering\small
\caption{Targets, required ADFs (circuits accepted by the implementation), and message tick count $T$ of the later circuit itself. Access to original inputs is described in Appendix~\ref{sec:tasks}.}
\label{tab:task-definitions}
\begin{tabularx}{\linewidth}{@{}>{\raggedright\arraybackslash}p{30mm}Y>{\raggedright\arraybackslash}p{20mm}c@{}}
\toprule
Task & Target formula & Required ADFs & $T$\\
\midrule
Has a father & $\exists y\,F(x,y)$ & None & 1\\
Has a sister & $\exists y\,S(x,y)$ & $S$ & 1\\
Grandparent & $\exists z[P(x,z)\wedge P(z,y)]$ & $P$ & 1\\
Uncle & $V(x,y)$ & $\operatorname{Bro}$ & 1\\
Maternal great-uncle & $\exists z[M(x,z)\wedge V(z,y)]$ & $V$ & 1\\
Adjacent to red & $\exists y[\operatorname{adj}(x,y)\wedge \operatorname{red}(y)]$ & None & 1\\
Reachability within 4 edges & $\operatorname{Reach}_2(x,y)\vee\exists z[\operatorname{Reach}_2(x,z)\wedge \operatorname{Reach}_2(z,y)]$ & $\operatorname{Reach}_2$ & 1\\
Reachability within 6 edges & $\operatorname{Reach}_4(x,y)\vee\exists z[\operatorname{Reach}_4(x,z)\wedge \operatorname{Reach}_2(z,y)]$ & \shortstack[l]{$\operatorname{Reach}_2$\\$\operatorname{Reach}_4$} & 1\\
Outdegree one & $|\{y:\operatorname{adj}(x,y)\}|=1$ & None & 1\\
Outdegree two & $|\{y:\operatorname{adj}(x,y)\}|=2$ & None & 1\\
\midrule
Sister relation & $S(x,y)$ & None & 2\\
Parent relation & $P(x,y)$ & None & 1\\
Brother relation & $\operatorname{Bro}(x,y)$ & None & 2\\
Reachability within 2 edges & $\operatorname{adj}(x,y)\vee\exists z[\operatorname{adj}(x,z)\wedge \operatorname{adj}(z,y)]$ & None & 2\\
\bottomrule
\end{tabularx}
\end{table}

For Has a sister, grandparent, and reachability within four or six edges, original unary and binary inputs are hidden from the later circuit after ADFs have been evaluated. Uncle combines the original parent relation with brother; Maternal great-uncle combines the original mother relation with uncle. Indirect dependencies, such as brother needed to evaluate the fixed uncle function, are supplied in a fixed order. The message ticks in the table exclude internal computation of already fixed ADFs.

Graphs are directed and have no self-edges. For each structure, an edge probability is drawn uniformly from $[0,0.3]$, and edges on ordered vertex pairs are generated independently. Each vertex is red with probability 0.25. Families assign sex with probability 0.5, then select father and mother candidates from people generated before the child. When candidates exist, one parent of each role is selected with probability 0.95, with male fathers and female mothers. Labels follow the formulas in Table~\ref{tab:task-definitions}. These probabilities specify the dataset distribution. The intended correctness statement quantifies over every structure in the declared domain.

Training and evaluation datasets contain distinct structures drawn under these assumptions (Table~\ref{tab:data}). Each dataset element is an entire structure. Every evaluated object or pair is labeled, yielding $n$ predictions per structure for unary tasks and $n(n-1)$ for binary tasks.
\begin{table}[!htb]
\centering\small
\caption{Numbers of structures at each object count $n$. Different uses have independent random-number streams.}
\label{tab:data}
\begin{tabularx}{\linewidth}{@{}>{\raggedright\arraybackslash}p{47mm}>{\raggedright\arraybackslash}p{43mm}Y@{}}
\toprule
Use & Object counts $n$ & Structures per count \\
\midrule
Training & 10, 20, 30, 40 & 256 each\\
Validation prediction report & 10, 20, 30, 40 & 128 each\\
Test prediction report & 10, 20, 30, 40 & 128 each\\
Tests with more objects & 50, 100, 200, 500 & 64, 32, 16, 4 respectively\\
Checks on the use of ADFs & 10, 20, 30, 40 & 128 each\\
Independent counterexample search & 50, 100, 200, 500 & 6 each\\
\bottomrule
\end{tabularx}
\end{table}
Validation and test predictions are for reporting only. They do not select weights, connections, initializations, training continuations, or saved circuits. Checks on the use of ADFs are part of acceptance and therefore use a separate set. Connection-removal checks use 16 additional fixed training structures at each training size.

\subsection{Training and evaluation protocol}
\label{sec:study}
Each of the fourteen tasks used seeds 0--9 to initialize ten acquisition attempts from scratch. Downstream tasks received previously acquired, frozen ADFs. Training, evaluation, and use checks had independent random-number streams. Each task ran all ten attempts. The first acquired program in seed order supplied its ADF to later tasks, so all downstream attempts using that dependency shared the same circuit. Tables~\ref{tab:primary-results} and~\ref{tab:aux-results} report the corresponding per-task intervals.

After 25,000 updates, the circuit is fixed and verified. If the implemented equivalence checks and required checks on the use of ADFs do not accept, the same weights, optimizer, and random state continue for another 25,000 updates. During this continuation, scheduled states with zero errors on the complete training dataset and at most six incoming edges per node are saved. After continuation finishes, at most four are checked in ascending lexicographic order of (maximum incoming edge count, mean incoming edge count, update number). The 50,000-update state is checked next if none is accepted. If it remains unaccepted but is correct on the complete training dataset, the post-training compression of Section~\ref{sec:method} is applied and the compressed circuit is checked. Checking stops at the first acceptance. Thus a trial invokes the checkpoint-checking pipeline at most seven times: one initial, up to four intermediate, one final, and one post-training compression check. Predictive evaluation does not determine this order. This rule is the same for all tasks and seeds.

Validation and test predictions are aggregated once for the selected checkpoint in every acquisition attempt. An unsuccessful attempt retains that checkpoint for diagnosis and returns no acquired program. A computing interruption resumes the saved weights, optimizer, and random state; it does not add a replacement seed. All 140 prescribed attempts completed, and every attempt invoked extraction and target-equivalence checking. The evidence links source and configuration hashes, dataset manifests, saved checkpoints, verification outcomes, and predictions; these links were checked before aggregation. Previously recorded experiments do not contribute to the reported denominators.

\subsection{Program acquisition}
\label{sec:results}
Neural NP4G acquired programs with proved target equivalence in 74 of 100 primary attempts (95\% binomial reference interval $[64.3,82.3]\%$) and 22 of 40 auxiliary attempts ($[38.5,70.7]\%$), under the semantics and input constraints of Section~\ref{sec:verification}. These fractions measure fixed-budget acquisition success across all prescribed attempts. The remaining 44 attempts returned no acquired program: the selected diagnostic candidates of 18 had counterexamples and those of 26 remained inconclusive, with 23 formulas outside the supported relational grammar and three outside the supported counting grammar. The recorded checkpoint-checking pipeline ran 294 times. There were 34 resource-limited checks in 28 distinct attempts: 11 time limits and 23 memory limits. The fixed continuation and selection procedure superseded all 34 intermediate resource outcomes. The check-count distribution was 69 attempts with one check, 52 with two, three with three, and 16 with seven. The primary and auxiliary denominators are reported separately because auxiliary tasks supply components of later programs.

Seed-trial proportions are accompanied by two-sided 95\% Clopper--Pearson intervals \citep{clopper1934limits}, with endpoints displayed in percent. Per-task intervals use a binomial model for initialization success conditional on the fixed dependency library. The pooled intervals summarize the prescribed attempts across heterogeneous tasks. For example, 2/10 gives $[2.5,55.6]\%$, 9/10 gives $[55.5,99.7]\%$, and even 10/10 gives $[69.2,100.0]\%$.

Every primary task yielded at least one acquired program with proved target equivalence (Table~\ref{tab:primary-results}), with acquisition success in two to ten of ten attempts per task. The floating-point validation and test evaluations of every acquired program also had zero errors.
\begin{table}[htbp]
\centering\small
\caption{Primary-task acquisition outcomes from ten attempts per task. Acquired counts returned programs, including required checks on the use of ADFs. Counterexample and inconclusive describe selected diagnostic candidates from attempts that returned no program. The three outcomes partition all attempts. Brackets give 95\% Clopper--Pearson intervals in percent for each outcome.}
\label{tab:primary-results}
\begin{tabularx}{\linewidth}{@{}YC{0.19\linewidth}C{0.19\linewidth}C{0.19\linewidth}@{}}
\toprule
Task & Acquired & Counterexample & Inconclusive\\
\midrule
Has a father & \trialrate{10}{10}{69.2}{100.0} & \trialrate{0}{10}{0.0}{30.8} & \trialrate{0}{10}{0.0}{30.8} \\
Has a sister & \trialrate{10}{10}{69.2}{100.0} & \trialrate{0}{10}{0.0}{30.8} & \trialrate{0}{10}{0.0}{30.8} \\
Grandparent & \trialrate{9}{10}{55.5}{99.7} & \trialrate{0}{10}{0.0}{30.8} & \trialrate{1}{10}{0.3}{44.5} \\
Uncle & \trialrate{4}{10}{12.2}{73.8} & \trialrate{2}{10}{2.5}{55.6} & \trialrate{4}{10}{12.2}{73.8} \\
Maternal great-uncle & \trialrate{2}{10}{2.5}{55.6} & \trialrate{3}{10}{6.7}{65.2} & \trialrate{5}{10}{18.7}{81.3} \\
Adjacent to red & \trialrate{10}{10}{69.2}{100.0} & \trialrate{0}{10}{0.0}{30.8} & \trialrate{0}{10}{0.0}{30.8} \\
Reachability within 4 edges & \trialrate{10}{10}{69.2}{100.0} & \trialrate{0}{10}{0.0}{30.8} & \trialrate{0}{10}{0.0}{30.8} \\
Reachability within 6 edges & \trialrate{2}{10}{2.5}{55.6} & \trialrate{8}{10}{44.4}{97.5} & \trialrate{0}{10}{0.0}{30.8} \\
Outdegree one & \trialrate{9}{10}{55.5}{99.7} & \trialrate{0}{10}{0.0}{30.8} & \trialrate{1}{10}{0.3}{44.5} \\
Outdegree two & \trialrate{8}{10}{44.4}{97.5} & \trialrate{0}{10}{0.0}{30.8} & \trialrate{2}{10}{2.5}{55.6} \\
\bottomrule
\end{tabularx}
\end{table}

Each trial evaluated 12,800 validation and 24,400 test predictions for unary tasks, or 371,200 validation and 2,479,600 test predictions for binary tasks. Tests included structures with up to 500 objects. There were 628 test structures per trial, of which only four had 500 objects. In binary tasks these four structures contribute $4\times500\times499=998{,}000$ labels, or 40.2\% of all test predictions. Labels within each structure share its relations, so pooling labels weights larger structures more heavily. Appendix~\ref{app:prediction-detail} derives bounds on the number of entirely correct structures.

\begin{table}[htbp]
\centering\small
\caption{The four auxiliary tasks, using the acquisition outcome definitions of Table~\ref{tab:primary-results}. Their 40 attempts are excluded from the primary denominator.}
\label{tab:aux-results}
\begin{tabularx}{\linewidth}{@{}YC{0.19\linewidth}C{0.19\linewidth}C{0.19\linewidth}@{}}
\toprule
Auxiliary task & Acquired & Counterexample & Inconclusive\\
\midrule
Sister relation & \trialrate{1}{10}{0.3}{44.5} & \trialrate{3}{10}{6.7}{65.2} & \trialrate{6}{10}{26.2}{87.8} \\
Parent relation & \trialrate{10}{10}{69.2}{100.0} & \trialrate{0}{10}{0.0}{30.8} & \trialrate{0}{10}{0.0}{30.8} \\
Brother relation & \trialrate{1}{10}{0.3}{44.5} & \trialrate{2}{10}{2.5}{55.6} & \trialrate{7}{10}{34.8}{93.3} \\
Reachability within 2 edges & \trialrate{10}{10}{69.2}{100.0} & \trialrate{0}{10}{0.0}{30.8} & \trialrate{0}{10}{0.0}{30.8} \\
\bottomrule
\end{tabularx}
\end{table}

All four auxiliary tasks yielded acquired programs (Table~\ref{tab:aux-results}). Seven programs were acquired after post-training compression: one sister, one brother, three grandparent, and two six-edge reachability programs. The selected ADFs were reused without changing their weights. In the six primary tasks requiring ADFs, all 37 acquired programs passed the specified formula-occurrence, active-gate, and individual-disabling checks. The checks establish use of each required dependency on the independent use-check datasets.

Figure~\ref{fig:acceptance-stages} shows the selection stages: 69 accepted initial checkpoints, 17 intermediate checkpoints, three final checkpoints, and seven post-training compressed checkpoints. Their primary/auxiliary splits are 52/17, 14/3, 3/0, and 5/2 respectively.
\input{generated/acceptance_stages_en}

Table~\ref{tab:finite-results} gives the finite prediction diagnostics of selected checkpoints. Among those with zero validation and test errors, five of 79 primary and seven of 29 auxiliary attempts had an inconclusive verification record (Table~\ref{tab:prediction-verification}). Appendix~\ref{sec:failure-analysis} analyzes examples from these saved formulas.
\begin{table}[htbp]
\centering\small
\caption{Diagnostic floating-point predictions from selected checkpoints of all acquisition attempts, including attempts that returned no program. The middle column counts attempts with no validation or test errors, with 95\% Clopper--Pearson intervals in brackets (\%). The final column pools test errors and object or pair labels over ten seeds. Asterisks mark auxiliary tasks.}
\label{tab:finite-results}
\begin{tabularx}{\linewidth}{@{}YC{0.20\linewidth}R{0.34\linewidth}@{}}
\toprule
Task & Zero-error trials & Test errors / predictions\\
\midrule
Has a father & \trialrate{10}{10}{69.2}{100.0} & 0 / 244,000 \\
Has a sister & \trialrate{10}{10}{69.2}{100.0} & 0 / 244,000 \\
Grandparent & \trialrate{9}{10}{55.5}{99.7} & 2,369,282 / 24,796,000 \\
Uncle & \trialrate{4}{10}{12.2}{73.8} & 7,332,078 / 24,796,000 \\
Maternal great-uncle & \trialrate{4}{10}{12.2}{73.8} & 2,882,704 / 24,796,000 \\
Adjacent to red & \trialrate{10}{10}{69.2}{100.0} & 0 / 244,000 \\
Reachability within 4 edges & \trialrate{10}{10}{69.2}{100.0} & 0 / 24,796,000 \\
Reachability within 6 edges & \trialrate{2}{10}{2.5}{55.6} & 604,804 / 24,796,000 \\
Outdegree one & \trialrate{10}{10}{69.2}{100.0} & 0 / 244,000 \\
Outdegree two & \trialrate{10}{10}{69.2}{100.0} & 0 / 244,000 \\
Sister relation$^{*}$ & \trialrate{6}{10}{26.2}{87.8} & 4,809,117 / 24,796,000 \\
Parent relation$^{*}$ & \trialrate{10}{10}{69.2}{100.0} & 0 / 24,796,000 \\
Brother relation$^{*}$ & \trialrate{3}{10}{6.7}{65.2} & 6,125,371 / 24,796,000 \\
Reachability within 2 edges$^{*}$ & \trialrate{10}{10}{69.2}{100.0} & 0 / 24,796,000 \\
\bottomrule
\end{tabularx}
\end{table}

\begin{table}[htbp]
\centering\small
\caption{Diagnostic prediction errors crossed with acquisition outcomes. Acquired counts returned programs; the other columns describe selected candidates from attempts that returned no program. ``Zero'' means no errors on either validation or test data; ``nonzero'' means at least one error on either dataset. Acquisition outcomes are the verdicts recorded at checkpoint selection.}
\label{tab:prediction-verification}
\begin{tabularx}{\linewidth}{@{}Ylrrrr@{}}
\toprule
Cohort & Prediction errors & Acquired & Counterexample & Inconclusive & Total\\
\midrule
Primary & Zero errors & 74 & 0 & 5 & 79 \\
Primary & Any error & 0 & 13 & 8 & 21 \\
Auxiliary & Zero errors & 22 & 0 & 7 & 29 \\
Auxiliary & Any error & 0 & 5 & 6 & 11 \\
\bottomrule
\end{tabularx}
\end{table}

Size-specific diagnostic accuracies and the distribution of test accuracy across seeds are reported in Appendix~\ref{app:prediction-detail}. For reachability within six edges, pooling predictions of selected candidates from all attempts yields 97.56\% accuracy, while proved acquisition succeeds in 2/10 attempts. Accuracy rises from 81.28\% at ten objects to 100\% at 500 objects, and all 40 of the 500-object test structures across ten seeds are predicted exactly. All 9,980,000 evaluated labels at this size are positive, so the constant-true predictor also gets all 40 structures right. The saved reports aggregate errors by object count. A zero-error group gives an exact count of entirely correct structures. For other groups, these aggregates generally give only bounds. Appendix~\ref{app:prediction-detail} reports these bounds.

A post hoc audit of the archived integer counts reports class prevalence and both constant predictors at 500 objects (Table~\ref{tab:label-balance}). A further baseline chooses the majority class separately for each seed using only its training labels, assigning ties to false. For four-edge reachability, constant true obtains 99.9589\%; for outdegree two, only 2 of 20,000 labels are positive, and constant false obtains 99.99\%. For several family tasks a constant exceeds the all-seed model accuracy at this size. The saved total error counts do not retain the class-specific errors needed to recover balanced accuracy or per-class recall.

\begin{table}[htbp]
\centering\scriptsize
\caption{Archived 500-object test labels, pooled over ten seeds (\%). ``Train-majority'' selects the constant from training labels only. Both fixed constants are shown without test-set selection. Auxiliary tasks carry an asterisk. Candidate accuracy includes checkpoints from attempts that returned no program. Four decimals expose severe class imbalance.}
\label{tab:label-balance}
\begin{tabularx}{\linewidth}{@{}Yrrrrr@{}}
\toprule
Task & Positive & False & True & Train-majority & Candidate\\
\midrule
Has a father & 94.7300 & 5.2700 & 94.7300 & 94.7300 & 100.0000\\
Has a sister & 69.8900 & 30.1100 & 69.8900 & 69.8900 & 100.0000\\
Grandparent & 0.6956 & 99.3044 & 0.6956 & 99.3044 & 90.1074\\
Uncle & 1.8265 & 98.1735 & 1.8265 & 98.1735 & 70.0057\\
Maternal great-uncle & 1.9844 & 98.0156 & 1.9844 & 98.0156 & 89.2072\\
Adjacent to red & 99.1750 & 0.8250 & 99.1750 & 99.1750 & 100.0000\\
Reachability within 4 edges & 99.9589 & 0.0411 & 99.9589 & 99.9589 & 100.0000\\
Reachability within 6 edges & 100.0000 & 0.0000 & 100.0000 & 100.0000 & 100.0000\\
Outdegree one & 2.5300 & 97.4700 & 2.5300 & 97.4700 & 100.0000\\
Outdegree two & 0.0100 & 99.9900 & 0.0100 & 99.9900 & 100.0000\\
Sister relation$^*$ & 0.3579 & 99.6421 & 0.3579 & 99.6421 & 80.2108\\
Parent relation$^*$ & 0.3786 & 99.6214 & 0.3786 & 99.6214 & 100.0000\\
Brother relation$^*$ & 0.7081 & 99.2919 & 0.7081 & 99.2919 & 75.3932\\
Reachability within 2 edges$^*$ & 84.2039 & 15.7961 & 84.2039 & 84.2039 & 100.0000\\
\bottomrule
\end{tabularx}
\end{table}

\subsection{What prevented acquisition}
\label{sec:failure-analysis}
All 26 final inconclusive records resolved the initial, intermediate, and final numerical channels and passed their finite circuit-versus-formula checks. Their first reported logical obstructions were negated binary atoms (14), negated existential joins (9), and formulas outside the single-relation counting grammar (3). Fourteen of their selected candidates also had validation or test prediction errors; twelve had none.

For example, the grandparent circuit from seed 7 extracted
\[
 (P(x,y)\wedge\neg P(y,x))\vee
 (\neg P(x,y)\wedge\neg P(y,x))
 \equiv\neg P(y,x),
\]
where $P$ is the proved parent ADF. Two unrelated objects satisfy this formula but not the grandparent target. The original checker stopped at the negated binary relation with an inconclusive verdict.

Two counting examples reveal errors missed by all recorded validation and test predictions. The outdegree-one circuit from seed 6 and outdegree-two circuit from seed 9 extracted, respectively,
\[
 \psi_k(x)=(\exists z\,\neg\operatorname{red}(z))
 \wedge [\min(\deg(x),17)=k],\qquad k\in\{1,2\}.
\]
Here $\deg(x)$ is adjacency outdegree. In a ten-object graph, color every object red, give object 0 exactly the edges $0\to1$ for $k=1$, or $0\to1$ and $0\to2$ for $k=2$, and include no other edges. These loop-free structures satisfy the declared input constraints. At object 0 the target is true and the extracted formula is false because the existential color condition fails. Evaluation of the saved extracted, closed, and domain-substituted formulas confirms these disagreements; CPU execution of the corresponding saved circuits also returns false. We constructed both counterexamples from the saved formulas after the experiment. They identify incorrect rules in two of the twelve zero-error trials originally recorded as inconclusive.

Conversely, the outdegree-two formula from seed 7 illustrates insufficient simplification. Write $n_{\mathrm{id}}=\min(|\{y:\operatorname{eye}(x,y)\}|,17)$ and $n_{\mathrm{deg}}=\min(\deg(x),17)$, where $\operatorname{eye}$ is identity. Its five disjuncts are exactly
\[
 \bigvee_{r=1}^{5}[n_{\mathrm{id}}\in\mathcal J_r\wedge n_{\mathrm{deg}}\in \mathcal D_r],\qquad
 \begin{array}{c|l|l}
 r&\mathcal J_r&\mathcal D_r\\\hline
 1&\{1,2,3,4\}&\{2\}\\
 2&\{5,6,7\}&\{2,3\}\\
 3&\{8,9,10,11\}&\{2,3,4\}\\
 4&\{12,13,14,15\}&\{2,3,4,5\}\\
 5&\{16,17\}&\{2,3,4,5,6\}
 \end{array}
\]
Identity has exactly one successor at every object, so $n_{\mathrm{id}}=1$ and only the first branch survives. The formula reduces to $\deg(x)=2$, the correct target. The archived checker encountered two count relations before eliminating the constant identity count. The identity substitution reduces this candidate to the supported single-relation fragment.

\subsection{Extracted programs}
\label{sec:formula-results}
The standard delivered formula applies target-independent, equivalence-preserving reduction before presentation. Identity aggregation is evaluated at its unique self-position, capped counts are reduced by their exact saturated truth profiles, and redundant conjuncts and disjuncts are removed only when containment justifies the deletion. Bound variables are renamed without capturing free arguments. The output uses the shorter reduced closed expression or a reduced body with named ADF calls and supplies each referenced definition once; binary outputs retain their false-self-pair guard. Raw extraction, source hashes, and the reduction record accompany this display.

Table~\ref{tab:compact-formulas} applies this policy to the fourteen selected archived circuits. Its closed and reduced columns compare the same expanded vocabulary and output guards; the standard column also permits named calls. The finite cross-check evaluated each of the fourteen formulas on 24 inputs, reusing the same inputs for formulas in the same domain. Across these 336 formula--structure pairs, the checks covered 2,416 in-domain arguments and 720 binary self-pairs with a required false output. On all 3,136 outputs the standard formula agreed with the archived formula obtained after dependency expansion and domain substitution. The equivalence justification comes from the recorded rewrite rules and containment checks.
\input{generated/compact_formulas_en.tex}
\input{generated/compact_examples_en.tex}

We measured the structure and character counts of saved extracted formulas. Counts include spaces, parentheses, and predicate names in the saved strings, before expanding the definitions of ADFs or applying any further simplification for readability. These character counts quantify the description length of the saved representations.

For adjacent-to-red, all ten trials were accepted by the verifier. Formula lengths ranged from 34 to 355,428 characters, with a median of 3,292. In mathematical notation, the seed 8 formula is $\rho(x)=\exists y[\operatorname{adj}(x,y)\wedge\operatorname{red}(y)]$. Using this same $\rho(x)$ as an abbreviation, the 104-character formula from seed 4 is
\[
 (\neg\operatorname{red}(x)\wedge \rho(x))\vee
 (\operatorname{red}(x)\wedge \rho(x)).
\]
The latter saved extraction retains an unnecessary case distinction, showing that logically equivalent conditions can have different description lengths. Appendix~\ref{app:worked-example} traces a saved Has a father circuit from its active connections and weights through outward interval evaluation, extraction, simplification, and target equivalence. The measured example complements the execution schematic in Figure~\ref{fig:program} with saved parameters and recomputed enclosures.

The first accepted circuit in the fixed order for reachability within six edges (seed 0) has a 136-character extracted formula. Replacing the names of its ADFs and its variables with this paper's notation gives
\[
 \bigl(\neg \operatorname{Reach}_4(x,y)\wedge
 \exists z[\operatorname{Reach}_2(x,z)\wedge \operatorname{Reach}_4(z,y)]\bigr)\vee \operatorname{Reach}_4(x,y).
\]
Reading $\operatorname{Reach}_2$ and $\operatorname{Reach}_4$ as already accepted partial programs makes the downstream composition compact to inspect. This count, however, excludes the definitions of those ADFs. Table~\ref{tab:formula-diagnostics} compares the saved extracted and closed formulas, counts syntax nodes with and without shared subexpressions, and includes the referenced definitions once in a named-library description. The 355,428-character adjacent-to-red expression has 31,766 syntax occurrences but only 287 distinct ordered subexpressions; the archived closing and Boolean-minimization stage reduces it to 34 characters. In fact, all ten adjacent-to-red trials have the same 34-character closed formula. Conversely, the six-edge expression becomes 14,809 characters after substituting its ADFs and applying the same recorded stage, while its body plus the selected dependency bodies totals 76,251 characters. The later domain-substitution stage reduces the six-edge expression to 732 characters. Character counts therefore depend strongly on representation and simplification stage.

\input{generated/formula_diagnostics_en}
\input{generated/formula_lengths_en}

These metrics quantify the saved formula stages. The named-library count includes each referenced body once, but excludes definition headers and implicit off-diagonal guards; the closed formula includes those guards. Syntax-tree operator counts and depth describe the logical structure of each expression. All stages and seed-level measurements are retained in the accompanying evidence.

\subsection{Published NLM results and methodological comparison}
\label{sec:nlm}
Table~\ref{tab:nlm-results} reproduces predictive accuracy from Table~1 and training graduation from Table~4 of \citet{dong2019nlm}. The original evaluation uses 1,000 randomly generated test instances and averages object or pair classification accuracy. NLM was trained at size 20 for family tasks and size 10 for graph tasks, then evaluated at the two sizes shown. Its family generator simulates timelines and marriage; its graphs are described as undirected and generated using spatial nearest neighbors. These distributions differ from our typed acyclic family inputs and directed graphs with a separately sampled edge probability for each graph.
\begin{table}[htbp]
\centering\small
\caption{Published NLM results \citep[Tables~1 and~4]{dong2019nlm}. Test accuracy applies at both listed sizes. Graduation counts seeds meeting the original training-loss criterion; brackets are 95\% Clopper--Pearson intervals (\%) computed here from those counts. }
\label{tab:nlm-results}
\begin{tabularx}{\linewidth}{@{}YcccC{0.19\linewidth}@{}}
\toprule
Task & Train $n$ & Test $n$ & Test accuracy & Graduation\\
\midrule
Has a father & 20 & 20, 100 & 100\%, 100\% & \trialrate{10}{10}{69.2}{100.0} \\
Has a sister & 20 & 20, 100 & 100\%, 100\% & \trialrate{10}{10}{69.2}{100.0} \\
Grandparent & 20 & 20, 100 & 100\%, 100\% & \trialrate{10}{10}{69.2}{100.0} \\
Uncle & 20 & 20, 100 & 100\%, 100\% & \trialrate{9}{10}{55.5}{99.7} \\
Maternal great-uncle & 20 & 20, 100 & 100\%, 100\% & \trialrate{2}{10}{2.5}{55.6} \\
Adjacent to red & 10 & 10, 50 & 100\%, 100\% & \trialrate{9}{10}{55.5}{99.7} \\
Reachability within 4 edges & 10 & 10, 50 & 100\%, 100\% & \trialrate{10}{10}{69.2}{100.0} \\
Reachability within 6 edges & 10 & 10, 50 & 100\%, 100\% & \trialrate{6}{10}{26.2}{87.8} \\
Outdegree one & 10 & 10, 50 & 100\%, 100\% & \trialrate{10}{10}{69.2}{100.0} \\
Outdegree two & 10 & 10, 50 & 100\%, 100\% & \trialrate{10}{10}{69.2}{100.0} \\
\bottomrule
\end{tabularx}
\end{table}

The original NLM training uses Adam with learning rate 0.005 and batch size four, task-dependent depths and breadths, and budgets of 50,000--200,000 training examples \citep[Appendices~A.1 and~B.2]{dong2019nlm}. In particular, outdegree two uses maximum predicate arity four. NLM's reported graduation is 9/10 for Uncle and 6/10 for six-edge reachability, higher than our acquisition counts of 4/10 and 2/10. NLM counts seeds reaching its training-loss criterion. Our trials count verified acquisitions under the distributions and budgets specified above. Section~\ref{sec:limitations} specifies the matched comparisons needed to attribute this gap.

NLM learns predicate combinations through shared numerical operations. Its empirical accuracy estimate uses sampled test structures, while its expressivity argument concerns realization of a restricted Horn-clause class \citep[Appendices~B.6 and~D]{dong2019nlm}. DLM replaces learned selections with discrete choices and fuzzy operations with Boolean operations to extract programs, and can reuse invented predicates in subsequent learning \citep[Sections~4.2.3 and~4.3]{zimmer2023dlm}. Neural NP4G establishes the meaning of each particular returned program through the chain from its saved real-arithmetic circuit to a faithful formula and from that formula to the supplied target. These equalities quantify over every declared finite structure and evaluated tuple, including all larger structures in the domain.
\FloatBarrier

\FloatBarrier
\section{Additional Discussion and Evidence Scope}
\label{app:discussion}
The saved Has a father circuit illustrates how its learned weights determine an extracted rule (Appendix~\ref{app:worked-example}). Figures~\ref{fig:acceptance-stages} and~\ref{fig:formula-lengths} report when the checks accepted programs and how formula length varied across trials.

Table~\ref{tab:representations} shows how readers can inspect a learned rule in each of three representations.

\begin{table}[htbp]
\centering\small
\caption{How readers can inspect acquired rules. NLM and DLM entries describe their published methods; our entry uses saved circuits and formulas.}
\label{tab:representations}
\begin{tabularx}{\linewidth}{@{}>{\raggedright\arraybackslash}p{23mm}YY@{}}
\toprule
Method & Representation of the acquired rule & What readers can inspect\\
\midrule
NLM & MLP weights and predicate tensors & Quantification and layer connections are explicit; learned predicate combinations are numerical MLP computations.\\
DLM & A program of selected logical operations and intermediate predicates & Conditions and predicate dependencies are symbolic. Appendix B.2 of the original paper gives learned formulas and simplification examples.\\
Neural NP4G & A weighted circuit with a faithful logical formula and definitions of acquired ADFs & Case distinctions and calls to ADFs are explicit. The acquisition procedure proves equivalence between the real-arithmetic circuit, its formula, and the target throughout the declared domain.\\
\bottomrule
\end{tabularx}
\end{table}

\subsection{Evidence availability}
\label{app:evidence-scope}
The distribution preserves aggregate ledgers, generated tables, hashes, and analysis code. A hash-checked source-history recovery reconstructed 21 runtime source files, the original collector, and 1,841 reports and manifests, but did not include the 393 referenced checkpoints. A separate local archive audit matched all 2,234 recorded artifacts, including those checkpoints; that archive is not included in the distributed source tree. At the source-availability audit, twelve recorded source entries matched the distributed sources and ten differed, so historical re-execution requires the archived source snapshot and recorded numerical environment. The class-prevalence and reporting audits use preserved records.
\FloatBarrier

\FloatBarrier
\section{Extended Related Work}
\label{app:related}
Inductive logic programming (ILP) acquires logical hypotheses from examples and background knowledge \citep{muggleton1991ilp,cropper2022ilp}. Popper alternates hypothesis generation, testing against examples, and learning constraints from failed hypotheses \citep{cropper2021popper}. This is a direct symbolic alternative to learning a numerical circuit and establishing its logical meaning afterwards. Neural LP makes relation-composition learning differentiable \citep{yang2017neurallp}, while differentiable ILP ($\partial$ILP) optimizes rules within supplied templates \citep{evans2018dilp}. Differentiable proving learns soft symbolic matching through a proof architecture \citep{rocktaschel2017proving}, and DeepProbLog combines neural predicates with probabilistic logic programs \citep{manhaeve2018deepproblog}.

NLM represents predicates involving different numbers of objects as tensors and organizes relational reasoning through expansion, permutation, and reduction. Multilayer perceptrons learn predicate combinations with parameters shared across objects \citep{dong2019nlm}. Its perfect-generalization results use empirical test accuracy, while its expressivity analysis gives realization of a restricted Horn-clause class \citep[Appendices~B.6 and~D]{dong2019nlm}. DLM directly extends this architecture, replacing its multilayer perceptron logic modules with differentiable predicate selection and logical operations that permit discrete program extraction \citep{zimmer2023dlm}. It also reuses predicates acquired during earlier training stages. Neural NP4G reuses the frozen numerical circuit and substitutes its verified definition when checking downstream compositions.

Logical characterizations of graph neural networks (GNNs) connect neighbor aggregation with graded modal and finite-variable counting logics \citep{barcelo2020logical,grohe2021logic}. Neural NP4G's existence, universal, and bounded-sum operators have corresponding existential, universal, and counting interpretations after state signs and numerical ranges are established. Characterizing the full architecture also requires accounting for pair states and ordered relation composition. \citet{salzer2023limits} establish undecidability for expressive message-passing graph classifiers over unbounded finite graphs and decidability for restricted settings; \citet{benedikt2024decidability} use logical characterizations to obtain both decidability and undecidability results for GNN classes. Here, finite state encodings, capped aggregation, and fixed execution lengths provide finite extraction cases, followed by equivalence checking on the declared fragment.

Learning connectivity with weights also has established precedents. Genetic programming uses ADFs for subprogram reuse \citep{koza1994gp2}, NEAT evolves neural topologies and weights \citep{stanley2002neat}, and differentiable logic gate networks learn Boolean-operation choices \citep{petersen2022dlogic}. Pruning and retraining learn sparse connections \citep{han2015connections}; stochastic gates support differentiable $L_0$ regularization \citep{louizos2018lzero}. Deep Rewiring changes connections during supervised learning under a strict connection budget and explicitly draws on structural plasticity \citep{bellec2018rewiring}. Neural NP4G uses deterministic forward binary gates with surrogate gradients, error-guarded pruning, and extraction-directed training.

NP4G composes programs by connecting nodes for supplied functions and stored data in a directed network \citep{hara2024np4g}. Connections determine which values reach each function, and acquired networks are reused as ADFs in subsequent learning phases. The reported experiment randomly selects nodes and connections to search for networks matching the training dataset. Neural NP4G represents the learned computation by weighted numerical units whose connections and parameters are optimized jointly, and stores acquired circuits as ADFs. This allows gradient-based training of the program's structure and numerical behavior. Binary gates select nodes and edges. Execution also differs: NP4G executes functions when their inputs are available, whereas Neural NP4G performs a prescribed number of synchronous numerical updates. Logical extraction and target-equivalence checking are applied to the learned circuits, with accepted circuits registered as ADFs.

Reusable computation also supports the accumulation of program libraries. DreamCoder alternates program search with learning reusable abstractions and a neural search guide \citep{ellis2021dreamcoder}. Neural NP4G shares the objective of acquiring reusable procedures with this work and DLM, while using selected nodes and edges of a weighted circuit as its internal representation. Its library of ADFs stores acquired circuits for composition in subsequent tasks.

Rule extraction from neural networks predates current neural program induction. Towell and Shavlik extract refined M-of-N rules from knowledge-based neural networks, following insertion and numerical refinement of symbolic knowledge \citep{towell1993rules}; \citet{andrews1995survey} survey extraction and refinement techniques. \citet{weiss2018automata} extract automata from trained recurrent networks using queries, counterexamples, and abstraction. We use faithfulness in the computational sense discussed by \citet{lipton2018mythos,jacovi2020faithfulness}. For Neural NP4G the extracted formula must match the saved circuit's real-arithmetic output at every declared input (Eq.~\eqref{eq:fidelity}).

Verifiable representations can also be learned by distillation. Programmatically Interpretable Reinforcement Learning (PIRL) searches a programmatic policy language using a neural policy as a guide \citep{verma2018pirl}; VIPER extracts decision-tree policies and verifies properties of the resulting trees \citep{bastani2018viper}. Verification-aware training provides a complementary connection: interval bound propagation \citep{gowal2018ibp} and differentiable abstract interpretation \citep{mirman2018diffai} optimize bounds used for robustness certification. The synthetic interval-range loss in Appendix~\ref{app:training} samples extraction ranges to encourage computations amenable to verification.

Abstract interpretation supplies foundations for numerical enclosure \citep{cousot1977abstract}, and Reluplex checks properties of neural computations over real inputs \citep{katz2017reluplex}. Query containment supplies the logical comparison \citep{chandra1977cq}. For queries with negation, \citet{levy1993updates} and \citet{wei2003negation} develop completion and canonical-database methods. Our checker explicitly enumerates unary truth assignments and variable identifications while restricting binary atoms to positive occurrences. The grammar and its preservation assumptions are stated in Appendix~\ref{sec:ucq}.

Learned reasoning now includes both strong task performance and formally checked outputs. Prompting and reinforcement learning improve language-model reasoning \citep{wei2022cot,guo2025deepseekr1}; AlphaGeometry combines neural proposals with symbolic deduction \citep{trinh2024alphageometry}; and AlphaProof learns proof search in Lean \citep{hubert2026alphaproof}. AlphaEvolve constructs algorithms using code proposals and automated evaluation \citep{novikov2025alphaevolve}, and AlphaProof Code verifies selected software security properties against human-reviewed specifications \citep{chaudhuri2026alphaproofcode}. Studies also identify condition-dependent robustness limits \citep{mirzadeh2025gsm,dziri2023faith} and analyze how learned skills can support compositional generalization \citep{yao2026compositional}. DafnyCOMP evaluates difficulties in constructing specifications across function boundaries \citep{xu2026dafnycomp}. In our relational experiments, supplied specifications check the learned numerical circuits and their compositions.

\input{generated/detector_protocol_en.tex}

\end{document}

%% file: common.tex
\usepackage{iftex}
\ifXeTeX\else
  \PackageError{neural-np4g-paper}{This manuscript requires XeLaTeX}{Use xelatex or tectonic.}
\fi

\ifdefined\JAPANESE
  \usepackage{fontspec}
  \defaultfontfeatures{Ligatures=TeX}
  \usepackage{xeCJK}
  \IfFontExistsTF{Yu Mincho}{
    \setCJKmainfont{Yu Mincho}[Scale=1,ItalicFont={Yu Mincho}]
  }{
    \IfFontExistsTF{YuMincho}{
      \setCJKmainfont{YuMincho}[Scale=1,ItalicFont={YuMincho}]
    }{
      \IfFontExistsTF{Noto Serif CJK JP}{
        \setCJKmainfont{Noto Serif CJK JP}[Scale=1,ItalicFont={Noto Serif CJK JP}]
      }{
        \setCJKmainfont{MS Mincho}[Scale=1,ItalicFont={MS Mincho}]
      }
    }
  }
  \IfFontExistsTF{Yu Gothic}{
    \setCJKsansfont{Yu Gothic}[Scale=1]
    \setCJKmonofont{Yu Gothic}[Scale=1]
  }{
    \IfFontExistsTF{YuGothic}{
      \setCJKsansfont{YuGothic}[Scale=1]
      \setCJKmonofont{YuGothic}[Scale=1]
    }{
      \IfFontExistsTF{Noto Sans CJK JP}{
        \setCJKsansfont{Noto Sans CJK JP}[Scale=1]
        \setCJKmonofont{Noto Sans CJK JP}[Scale=1]
      }{
        \setCJKsansfont{MS Gothic}[Scale=1]
        \setCJKmonofont{MS Gothic}[Scale=1]
      }
    }
  }
  \XeTeXlinebreaklocale "ja"
\fi

\usepackage{microtype}
\usepackage{amsmath,amssymb,mathtools,bm}
\usepackage{amsthm}
\theoremstyle{definition}
\usepackage{booktabs,tabularx,array,multirow}
\usepackage{longtable}
\usepackage{siunitx}
\usepackage{xcolor}
\usepackage{graphicx}
\usepackage{tikz}
\usetikzlibrary{arrows.meta,positioning,fit,calc,backgrounds,shapes.geometric}
\usepackage{enumitem}
\usepackage{float}
\usepackage{placeins}
\usepackage{url}
\usepackage[hidelinks,pdfencoding=auto,psdextra]{hyperref}

\definecolor{NPBlue}{HTML}{1E5AA8}
\definecolor{NPTeal}{HTML}{16827A}
\definecolor{NPGold}{HTML}{C98611}
\definecolor{NPRed}{HTML}{B33A3A}
\definecolor{NPPurple}{HTML}{6A4C93}
\definecolor{NPLightBlue}{HTML}{EAF2FB}
\definecolor{NPLightTeal}{HTML}{E7F4F2}
\definecolor{NPLightGold}{HTML}{FFF5DD}
\definecolor{NPLightRed}{HTML}{FBEAEA}
\definecolor{NPGray}{HTML}{5A6573}
\definecolor{NPLightGray}{HTML}{F2F4F7}

\hypersetup{
  colorlinks=true,
  linkcolor=NPBlue,
  citecolor=NPTeal,
  urlcolor=NPPurple
}

\newcolumntype{Y}{>{\raggedright\arraybackslash}X}
\newcolumntype{C}[1]{>{\centering\arraybackslash}p{#1}}
\newcolumntype{R}[1]{>{\raggedleft\arraybackslash}p{#1}}


\newcommand{\relu}{\operatorname{ReLU}}
\newcommand{\sg}{\operatorname{stopgrad}}
\newcommand{\Gclass}{\mathcal{G}}

\newcommand{\Circuit}{\mathcal{C}}

\newcommand{\trialrate}[4]{\shortstack{#1/#2\\[-1pt]{\scriptsize $[#3,#4]\%$}}}

\newcommand{\resultmark}[1]{%
  \ifstrequal{#1}{yes}{\textcolor{NPTeal}{\ensuremath{\bullet}}}{\textcolor{NPRed}{\ensuremath{\circ}}}%
}

\usepackage{xstring}

\providecommand{\PaperPDFAuthor}{Anonymous authors}
\providecommand{\PaperPDFCreator}{Anonymous review}
\providecommand{\PaperPDFProducer}{Anonymous review}
\AtBeginDocument{%
  \hypersetup{%
    pdftitle={\PaperPDFTitle},pdfauthor={\PaperPDFAuthor},%
    pdfcreator={\PaperPDFCreator},pdfproducer={\PaperPDFProducer},%
    pdfcreationdate={D:19700101000000Z},pdfmoddate={D:19700101000000Z}%
  }%
}

%% file: authors_en.tex
\author{\name Shoichiro Hara \email s.hara@nsc.nagoya-cu.ac.jp \\
        \addr Nagoya City University
        \AND
        \name Yuji Watanabe \email yuji@nsc.nagoya-cu.ac.jp \\
        \addr Nagoya City University}
\renewcommand{\PaperPDFAuthor}{Shoichiro Hara and Yuji Watanabe}
\renewcommand{\PaperPDFCreator}{LaTeX}
\renewcommand{\PaperPDFProducer}{LaTeX}

%% file: generated/relational_architecture.tex
% Shared architecture figure; labels follow the manuscript language.
\begin{figure}[!t]
\centering
\begingroup
\ifdefined\JAPANESE
 \def\RawLabel{入力構造 $X^{(0)},X^{(1)},X^{(2)}$}
 \def\ADFLabel{獲得済み ADFs\\依存順に1回だけ評価}
 \def\AugLabel{出力を入力特徴へ結合\\$\bar X^{(1)},\bar X^{(2)}$}
 \def\UnaryLabel{単項課題：対象ごとの状態}
 \def\BinaryLabel{二項課題：対象対ごとの状態}
 \def\OpsLabel{基本演算と選択ゲート}
 \def\InitLabel{初期化}
 \def\CallLabel{共有局所回路\\$K$ circuit ticks\\内部値を毎回リセット}
 \def\RepeatLabel{$t=0,\ldots,T-1$：各行のメッセージティックを反復}
 \def\ReadLabel{$T$ 回後：線形読出し $\to\ \mathbf 1[\mathrm{logit}\geq0]$}
\else
 \def\RawLabel{Input structure $X^{(0)},X^{(1)},X^{(2)}$}
 \def\ADFLabel{Acquired ADFs\\evaluate once in dependency order}
 \def\AugLabel{Concatenate outputs with inputs\\$\bar X^{(1)},\bar X^{(2)}$}
 \def\UnaryLabel{Unary task: object states}
 \def\BinaryLabel{Binary task: pair states}
 \def\OpsLabel{Primitive operators and gates}
 \def\InitLabel{Initialize}
 \def\CallLabel{Shared local circuit\\$K$ circuit ticks\\reset internal values per call}
 \def\RepeatLabel{$t=0,\ldots,T-1$: repeat the message tick in the applicable row}
 \def\ReadLabel{After $T$ ticks: linear readout $\to\ \mathbf 1[\mathrm{logit}\geq0]$}
\fi
\resizebox{\linewidth}{!}{%
\begin{tikzpicture}[
 b/.style={draw=NPBlue,fill=NPLightBlue,rounded corners=2pt,align=center,minimum height=10mm,font=\small},
 p/.style={b,draw=NPTeal,fill=NPLightTeal},
 a/.style={-{Latex[length=2mm]},thick,draw=NPGray},
 lab/.style={font=\small,align=center}]
 \node[b,text width=40mm] (raw) at (1.8,0) {\RawLabel};
 \node[p,text width=44mm] (adf) at (7.3,0) {\ADFLabel};
 \node[b,text width=44mm] (aug) at (12.8,0) {\AugLabel};
 \draw[a] (raw)--(adf);
 \draw[a] (adf)--(aug);
 \draw[a] (raw.north)--++(0,.55)-|(aug.north);
 \node[lab] (init) at (7.3,-1.35)
   {\InitLabel: $S_i^0=h(L_n(\bar X_i^{(1)})+L_g(X^{(0)}))$\quad
    $\mathsf Q_{ij}^0=h(L_e(\bar X_{ij}^{(2)})+L_s(\bar X_i^{(1)})+L_t(\bar X_j^{(1)}))$};
 \draw[a] (aug.south)--++(0,-.3)-|(init.north);
 \node[lab,text=NPBlue,anchor=west] at (-.3,-2.25) {\UnaryLabel};
 \node[b,text width=15mm] (s) at (.5,-3.2) {$S^t$};
 \node[p,text width=39mm] (unary) at (3.8,-3.2)
   {$E,U,N$ (gated); $\max_j S_j^t$\\$R\in\bar X^{(2)}$: $R$ and $R^\top$};
 \node[b,text width=18mm] (ucat) at (7.1,-3.2) {$[S^t;E;U;$\\$N;\max]$};
 \node[b,text width=35mm] (uf) at (10.5,-3.2) {$F_\theta$\\\CallLabel};
 \node[p,text width=15mm] (uh) at (13.7,-3.2) {$h$\\$S^{t+1}$};
 \draw[a] (s)--(unary); \draw[a] (unary)--(ucat);
 \draw[a] (ucat)--(uf); \draw[a] (uf)--(uh);
 \draw[a] (s.south)--++(0,-.7)-|(ucat.south);
 \draw[a] (uh.south)--++(0,-1.1)-|(s.south);
 \node[lab,text=NPBlue,anchor=west] at (-.3,-5.1) {\BinaryLabel};
 \node[b,text width=15mm] (q) at (.5,-6.05) {$\mathsf Q^t$};
 \node[p,text width=39mm] (join) at (3.8,-6.05)
   {$J^{cd}_{ij}(\mathsf Q^t)$ (gated)\\$\max_k$: relation composition};
 \node[b,text width=18mm] (bcat) at (7.1,-6.05) {$[\mathsf Q^t;(\mathsf Q^t)^\top;$\\$J;R;R^\top]$};
 \node[b,text width=35mm] (bf) at (10.5,-6.05) {$F_\theta$\\\CallLabel};
 \node[p,text width=15mm] (bh) at (13.7,-6.05) {$h$\\$\mathsf Q^{t+1}$};
 \draw[a] (q)--(join); \draw[a] (join)--(bcat);
 \draw[a] (bcat)--(bf); \draw[a] (bf)--(bh);
 \draw[a] (q.south)--++(0,-.7)-|(bcat.south);
 \draw[a] (bh.south)--++(0,-1.1)-|(q.south);
 \node[lab,anchor=south] at (7.1,-5.25) {$R\in\bar X^{(2)}$};
 \node[lab] at (7.3,-8.3) {\RepeatLabel};
 \node[b,text width=135mm] at (7.3,-9.1) {\ReadLabel};
 \draw[a] (init.west)-|(-.65,-3.2)--(s.west);
 \draw[a] (-.65,-3.2)--(-.65,-6.05)--(q.west);
\end{tikzpicture}}
\ifdefined\JAPANESE
\caption{関係計算の全体像。上段で獲得済み ADF の出力を元の入力へ結合する。二項 ADF の自己対は偽に設定する。課題の出力次数に応じて単項または二項の行を実行する。基本演算は各メッセージティックで現在の状態に作用し、結合した特徴を共有回路 $F_\theta$ へ渡す。$h$ は初期化時と各回路呼出し直後に適用され、反復間では二値状態を保持する。学習時の連続緩和は本文の $h_\alpha$ に従う。}
\else
\caption{Complete relational computation. Acquired ADF outputs are concatenated with the original inputs once; binary ADF self-pairs are set to false. The target arity selects the unary or binary row. Primitive operators act on the current states at every message tick, and concatenated features enter the shared circuit $F_\theta$. Thresholding $h$ occurs at initialization and immediately after each circuit call; binary states persist between ticks. Training uses the relaxation $h_\alpha$ defined in the text.}
\fi
\label{fig:relational-architecture}
\endgroup
\end{figure}

%% file: generated/detector_experiment_en.tex
\subsection{Detector information and counterexample feedback}
\label{sec:detector-experiment}
We tested edge reversal and edge symmetrization on directed loop-free graphs, evaluating binary outputs at $x\ne y$. Ten matched seeds, four detector settings, and feedback disabled or enabled give 160 neural training runs. Each uses one state channel, three hidden nodes, $K=2$, $T=1$, no composition primitive, and one randomly generated three-object labeled graph. The budget is 600 optimizer updates, with a detector query every 200 updates; acceptance stops the trajectory.

The four detectors are: \textbf{(a) Formula}, which exposes the target formula and checks exact equivalence; \textbf{(b) Black box}, which returns the same decision and one labeled counterexample while withholding the formula; \textbf{(c) Partial}, which completely checks edge monotonicity and, for symmetrization, output symmetry within the local formula fragment; and \textbf{(d) Finite}, which accepts after finding no disagreement on eight fixed four-object graphs. Both constant rules satisfy (c), making its specification strictly weaker than the target. Conditions (a) and (b) retain the same semantic information: their common optimizer does not read the formula.

For distinct vertices $x,y$, a local formula depends only on $\operatorname{adj}(x,y)$, $\operatorname{adj}(y,x)$, $\operatorname{red}(x)$, and $\operatorname{red}(y)$. Each of the 16 assignments occurs on two vertices, and every distinct pair in a graph of any size induces one of them. Agreement on this truth table therefore proves equivalence of the extracted and target formulas in the local fragment.

Feedback retains the original supervised loss and adds a cyclically selected counterexample loss. Labeled feedback supervises one returned output cell; property feedback penalizes a symmetry or monotonicity violation without supplying a target label. After stopping, an independent audit re-extracted all 160 saved circuits and checked their formulas against the targets. It also compared circuit and formula outputs on the 16 assignments in both vertex orders. All $160\times16\times2=5{,}120$ output cells agreed. The audit was withheld from training (Appendix~\ref{app:detector-protocol}).

\input{generated/detector_study_en.tex}

Exact-detector feedback increases verified acquisition from 5/10 to 8/10 on reversal and from 2/10 to 5/10 on symmetrization (Table~\ref{tab:detector-study}). Formula and black-box conditions have identical paired trajectories. With feedback, partial specifications accept all ten runs per task but retain four and eight incorrect rules, respectively. Finite search obtains 8/10 and 7/10 correct accepted programs, while accepting two incorrect reversal rules. Its symmetrization acquisition of 7/10 exceeds the exact detector's 5/10: counterexample choice and the finite training budget affect acquisition, while checking strength controls acceptance reliability.

%% file: generated/detector_study_en.tex
% Generated from docs/detector_study_20261115/neural_results.json after independent audit.
\begin{table}[htbp]
\centering
\small
\setlength{\tabcolsep}{5pt}
\caption{Bounded neural detector study: each entry gives replay disabled $\to$ enabled, with ten paired seeds per condition. Accepted counts detector acceptance; acquired additionally requires independent exact target equivalence. Mean queries includes all ten runs, including budget exhaustion and erroneous acceptance. The budget is 600 optimizer updates with a query every 200 updates.}
\label{tab:detector-study}
\begin{tabular}{llrrr}
\toprule
Task & Detector & Accepted & Acquired & Mean queries \\
\midrule
Edge reversal & Formula & $5\to 8$ & $5\to 8$ & $2.0\to 1.9$ \\
 & Black box & $5\to 8$ & $5\to 8$ & $2.0\to 1.9$ \\
 & Partial & $8\to 10$ & $5\to 6$ & $1.4\to 1.2$ \\
 & Finite & $6\to 10$ & $5\to 8$ & $1.8\to 1.5$ \\
\midrule
Edge symmetrization & Formula & $2\to 5$ & $2\to 5$ & $2.6\to 2.4$ \\
 & Black box & $2\to 5$ & $2\to 5$ & $2.6\to 2.4$ \\
 & Partial & $8\to 10$ & $2\to 2$ & $1.4\to 1.3$ \\
 & Finite & $2\to 7$ & $2\to 7$ & $2.6\to 2.4$ \\
\bottomrule
\end{tabular}
\end{table}

%% file: generated/method_details_en.tex
\subsection{Execution examples and the two-tick reduction}
\label{app:execution-details}
\begin{figure}[htbp]
\centering
\begin{tikzpicture}[
 unit/.style={circle,draw=NPBlue,fill=NPLightBlue,minimum size=10mm,font=\small},
 arr/.style={-{Latex[length=2mm]},thick,draw=NPBlue}]
 \node[unit] (x) at (0,0) {$x$};
 \node[unit] (h) at (2,0) {$h$};
 \node[unit] (o) at (4,0) {$o$};
 \draw[arr] (x)--node[above]{$w_{xh}$}(h);
 \draw[arr] (h)--node[above]{$w_{ho}$}(o);
 \draw[dashed,draw=NPGray,-{Latex}] (x) to[bend right=35]
 node[below,font=\small]{$g_{xo}=0$}(o);
 \node[align=left,anchor=west,font=\small] at (5.1,0) {
 tick 0: $h=0,\ o=0$\\
 tick 1: $h=\sigma(w_{xh}x+b_h),\ o=b_o$\\
 tick 2: $o=w_{ho}\sigma(w_{xh}x+b_h)+b_o$};
\end{tikzpicture}
\caption{Schematic composition through connections and ticks. The two solid edges are selected; the dashed direct connection is disabled. The hidden and output biases are $b_h$ and $b_o$. The schematic illustrates circuit execution.}
\label{fig:program}
\end{figure}
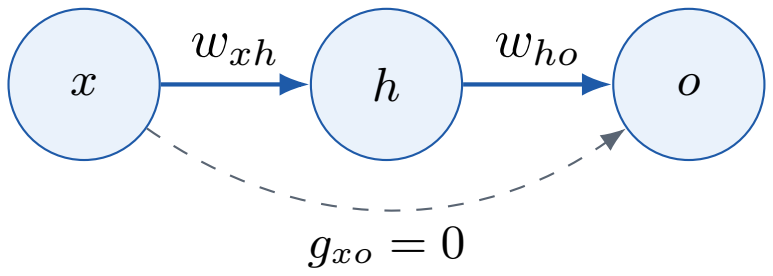

In the two-edge path of Figure~\ref{fig:program}, the hidden node processes the input at tick one and passes its result to the output at tick two. Selecting a direct input-to-output connection or a path through a hidden node changes how computations compose. At $K=2$, however, hidden-to-hidden and output-to-hidden connections cannot affect the final output; their effects require at least three ticks to reach it. Output-to-output connections can already contribute affine terms at two ticks. The complete model combines this local circuit with aggregation, thresholded states, relational updates, and reuse of ADFs as described below.

The effective local function at two ticks can be stated more precisely. Let $A_{UV}$ denote the effective weight block with destination $U$ and source $V$, and let $D_{\mathcal H}=\operatorname{diag}(g_{\mathcal H})$. With $v_{\mathrm{aff}}=b_O+A_{OI}x$ and $v_{\mathrm{hid}}=D_{\mathcal H}\sigma(b_H+A_{HI}x)$, zero initialization gives
\begin{equation}
 F_\theta(x)=(I_{\mathrm{out}}+A_{OO})v_{\mathrm{aff}}+A_{OH}v_{\mathrm{hid}}.
 \label{eq:two-tick-reduction}
\end{equation}
Thus each frozen local circuit is representable by a single sigmoid hidden layer with an affine input skip. Gating, pruning, and factored affine parameters can still affect optimization, and the outer relational computation remains. A detached conversion and numerical equivalence checks accompany this identity. Its implications for comparisons and floating-point execution are discussed in Section~\ref{sec:limitations}.

\subsection{Gate initialization and compression}
Initialization deliberately favors input--hidden--output paths. Input-to-hidden and hidden-to-output gate parameters start at $+2$; input-to-output and hidden-to-hidden parameters start at zero, giving $p_e=0.5$ and therefore $g_e=0$ under the strict inequality. Other edge parameters start at $-2$, and hidden-node parameters at $+2$. A zero-initialized edge is initially inactive, while its gate parameter can still receive a surrogate gradient. Appendix~\ref{app:training} lists the remaining settings.

Connections are removed gradually. Reducing the number of incoming edges per node reduces the combinations considered during extraction. A removal is kept only when it preserves a currently correct training minibatch and does not increase error on an independent training check set. Once states are fully binary, a saved state with zero errors on the complete training dataset can be restored if subsequent updates break this property. Appendix~\ref{app:training} gives the losses and numerical settings.

If a circuit remains unaccepted after training but has zero errors on the complete training dataset, a further compression step is applied. Hidden nodes are disabled in index order, followed by edges in increasing absolute effective-weight order, retaining only changes that preserve correctness on the complete training dataset. At most 128 edges are tried. Predictive evaluation data and the target formula are unavailable to these decisions. The compressed circuit is then extracted and verified again. Appendix~\ref{sec:study} specifies its place in the fixed training procedure.
\FloatBarrier

\subsection{Composition and the acquisition pipeline}
For example, given a relation $R(x,y)$, composing two instances of that relation gives
\begin{equation}
 T(x,y)=\exists z\,[R(x,z)\wedge R(z,y)].
 \label{eq:example}
\end{equation}
For a directed edge relation $R$, this target expresses reachability through two edges. The target rule specifies the labels; learning acquires the circuit connections and weights that implement that rule.

As an example of reuse, consider learning a new rule from a single acquired ADF with definition $\psi_1$, including the relation obtained by composing it with itself:
\begin{equation}
 \psi_2(x,y)=\psi_1(x,y)\vee\exists z[\psi_1(x,z)\wedge\psi_1(z,y)].
 \label{eq:adf-compose}
\end{equation}
The downstream circuit receives this ADF's outputs and learns which components and composition results to use. Once verification accepts the new circuit, it is also registered as an ADF. Acquired partial programs thus become components of subsequent structure learning.

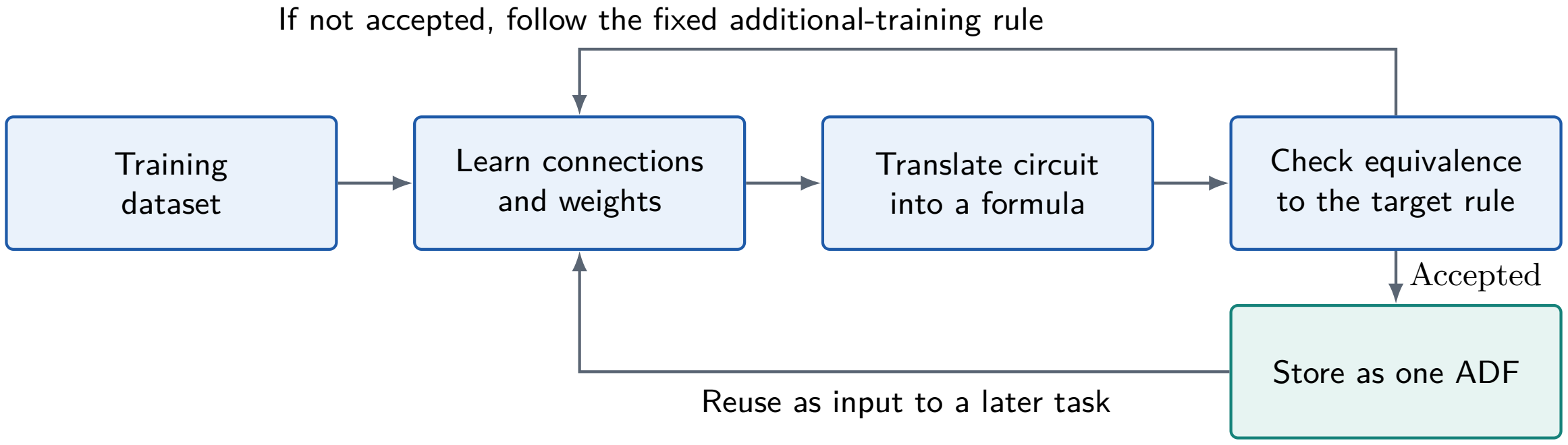
\begin{figure}[htbp]
\centering
\begin{tikzpicture}[
 box/.style={rounded corners=2pt,draw=NPBlue,fill=NPLightBlue,thick,
 align=center,text width=30mm,minimum height=13mm,font=\small\sffamily},
 arr/.style={-{Latex[length=2mm]},thick,draw=NPGray}]
 \node[box] (data) at (0,0) {Training\\dataset};
 \node[box] (learn) at (4,0) {Learn connections\\and weights};
 \node[box] (formula) at (8,0) {Translate circuit\\into a formula};
 \node[box] (proof) at (12,0) {Check equivalence\\to the target rule};
 \draw[arr] (data)--(learn);
 \draw[arr] (learn)--(formula);
 \draw[arr] (formula)--(proof);
 \draw[arr] (proof.north) -- ++(0,0.65) -|
 node[pos=0.45,above,font=\small\sffamily]{If not accepted, follow the fixed additional-training rule}(learn.north);
 \node[box,fill=NPLightTeal,draw=NPTeal] (adf) at (12,-1.85) {Store as one ADF};
 \draw[arr] (proof.south) -- node[right,font=\small]{Accepted}(adf.north);
 \draw[arr] (adf.west) -- (4,-1.85) -- (learn.south);
 \node[font=\small\sffamily,anchor=north] at (7.2,-1.9) {Reuse as input to a later task};
\end{tikzpicture}
\caption{The evaluated acquisition procedure for Neural NP4G. Learning proposes numerical circuits; faithful extraction and target-equivalence proof determine which program is returned and stored as an ADF. Rejected candidates follow the fixed continuation protocol. Accepted ADFs supply input features to later tasks.}
\label{fig:pipeline}
\end{figure}

\subsection{Batched feature layout}
For a batch of $n_{\mathrm b}$ structures with at most $n_{\max}$ objects, global, per-object, and per-pair features have shapes
\[
 X^{(0)}\in\mathbb R^{n_{\mathrm b}\times f_0},\quad
 X^{(1)}\in\mathbb R^{n_{\mathrm b}\times n_{\max}\times f_1},\quad
 X^{(2)}\in\mathbb R^{n_{\mathrm b}\times n_{\max}\times n_{\max}\times f_2}.
\]
The feature counts are $f_0,f_1,f_2$; masks exclude padding from every aggregation. The gate estimator differentiates through $p_e$ while executing $g_e$, a generally biased surrogate gradient. Input and output nodes remain active throughout training.

%% file: generated/worked_example_en.tex
% Generated by paper/analyze_worked_example.py from an actual selected checkpoint.
\section{A measured circuit-to-formula trace}
\label{app:worked-example}
\label{sec:measured-trace}
\input{generated/measured_circuit_en}
This example uses the selected Has a father checkpoint from seed~1 after
41,000 updates, chosen for exposition because its output cone is small.
It is distinct from the seed-0 circuit registered in the library.
There are 55 active edges and 12 active hidden nodes in the saved circuit;
only seven hidden nodes can affect its four outputs within the two circuit
ticks.  The execution has one message tick and uses no learned ADFs.
The supplementary record identifies the checkpoint and extraction by full
SHA-256 hashes and provides every effective edge and interval endpoint,
including the saved dense embedding weights and biases.

Outward evaluation of the saved embedding encloses all its preactivations
in $[7.9506,9.2487]$, yielding $s_c^{(0)}(x)=+1$ for all four channels.
Thus the four father-aggregation inputs coincide:
\begin{equation}
v_{\rm agg}(x)=\begin{cases}
0,&\neg E(x),\\
\sigma(4)-\tfrac12,&E(x),
\end{cases}
\qquad E(x)\equiv\exists y\,\mathrm{father}(x,y).
\end{equation}
The nonzero value is enclosed by $[0.4820137900379083,
0.4820137900379088]$.  Substituting only these equal inputs and proven
constants into the saved connections gives the complete output cone
\begin{equation}
\begin{aligned}
h_k&=\sigma(\eta_k v_{\rm agg}+b_k),&
z_i&=\delta_i v_{\rm agg}+\epsilon_i+\sum_{k=1}^{7}W_{ik}h_k,\\
s_i&=\begin{cases}+1&z_i\geq0,\\-1&z_i<0,\end{cases}&
\widehat y&=\mathbf1[\ell\geq0].
\end{aligned}
\end{equation}
Here $h_k$ is the first-circuit-tick hidden value and $z_i$ is the
second-circuit-tick output; the affine readout is
\begin{equation}
\ell=\omega^{\mathsf T}s+\beta.
\end{equation}
Table~\ref{tab:measured-weights} lists the retained weights after this
exact substitution, with $\eta_k$ summing connections from the identical
aggregation inputs and $b_k$ including constant-input connections.
The decimals in that table and the coefficient vectors below denote
centres of enclosing intervals $[r-0.00005,r+0.00005]$; zero entries
are exactly zero.  The interval trace uses the original saved binary
parameters.
\begin{table}[H]
\centering\small
\caption{Measured coefficients of the complete output cone.}
\label{tab:measured-weights}
\begin{tabular}{rrrrrrr}
\toprule
$k$ & $\eta_k$ & $b_k$ & $W_{1k}$ & $W_{2k}$ & $W_{3k}$ & $W_{4k}$\\
\midrule
1 & 27.3054 & -0.4212 & 0.0000 & 0.0000 & -4.6449 & -5.9999 \\
2 & 7.4892 & 4.0135 & 2.2787 & 0.0000 & 0.0000 & 0.0000 \\
3 & 10.8905 & 4.0458 & 2.2452 & 0.0000 & 0.0000 & 0.0000 \\
4 & -24.1090 & -0.0916 & 0.0000 & 0.0000 & 1.5266 & 7.5269 \\
5 & -11.9579 & -4.0019 & 0.0000 & 0.0000 & 0.0000 & 3.4499 \\
6 & 11.9223 & 4.7061 & 2.0246 & 2.5054 & 0.0000 & 0.0000 \\
7 & 12.9534 & 5.6328 & 0.0000 & 2.2526 & 0.0000 & 0.0000 \\
\bottomrule
\end{tabular}
\end{table}
\begin{align}
\delta&=(0.0000,4.7525,-9.2896,0.0000),\\
\epsilon&=(1.1842,-1.1305,1.1168,-1.1235),\\
\omega&=(2.3206,2.3226,-4.7865,-7.1275),\qquad \beta=-1.1646.
\end{align}
The full outward trace determines every channel sign and the readout
in both cases (Table~\ref{tab:measured-intervals}); the displayed interval
endpoints are rounded outwards.
\begin{table}[H]
\centering\small
\caption{Outward intervals from the saved circuit, including the first
hidden node.  The last column is the resulting state $s_i$ for $z_i$ and
prediction for $\ell$; hidden activations are not discretized.}
\label{tab:measured-intervals}
\begin{tabular}{clrc}
\toprule
$E(x)$ & Quantity & Enclosing interval & Discrete value\\
\midrule
0 & $h_1$ & $[0.39623236,0.39623237]$ & --- \\
0 & $z_1$ & $[7.6355,7.6356]$ & $+1$ \\
0 & $z_2$ & $[3.5970,3.5971]$ & $+1$ \\
0 & $z_3$ & $[0.0047,0.0048]$ & $+1$ \\
0 & $z_4$ & $[0.1522,0.1523]$ & $+1$ \\
0 & $\ell$ & $[-8.4355,-8.4354]$ & $0$ \\
1 & $h_1$ & $[0.99999706,0.99999707]$ & --- \\
1 & $z_1$ & $[7.7314,7.7315]$ & $+1$ \\
1 & $z_2$ & $[5.9181,5.9182]$ & $+1$ \\
1 & $z_3$ & $[-8.0058,-8.0057]$ & $-1$ \\
1 & $z_4$ & $[-7.1232,-7.1231]$ & $-1$ \\
1 & $\ell$ & $[15.3924,15.3925]$ & $1$ \\
\bottomrule
\end{tabular}
\end{table}
In particular, the prediction is false
when no father exists and true otherwise, for any finite admissible
structure.  The recorded extraction is
\begin{equation}
\phi_{\rm extracted}(x)=\exists y\,[\mathrm{father}(x,y)\land\top].
\end{equation}
Removing the conjunction with $\top$ gives
$\phi_{\rm simplified}(x)=\exists y\,\mathrm{father}(x,y)$,
which equals the target.  Replaying extraction from this checkpoint
reproduces both the stored formula and the equivalence verdict.
This chain concerns the exact-real interpretation of the saved binary
parameters.

%% file: generated/measured_circuit_en.tex
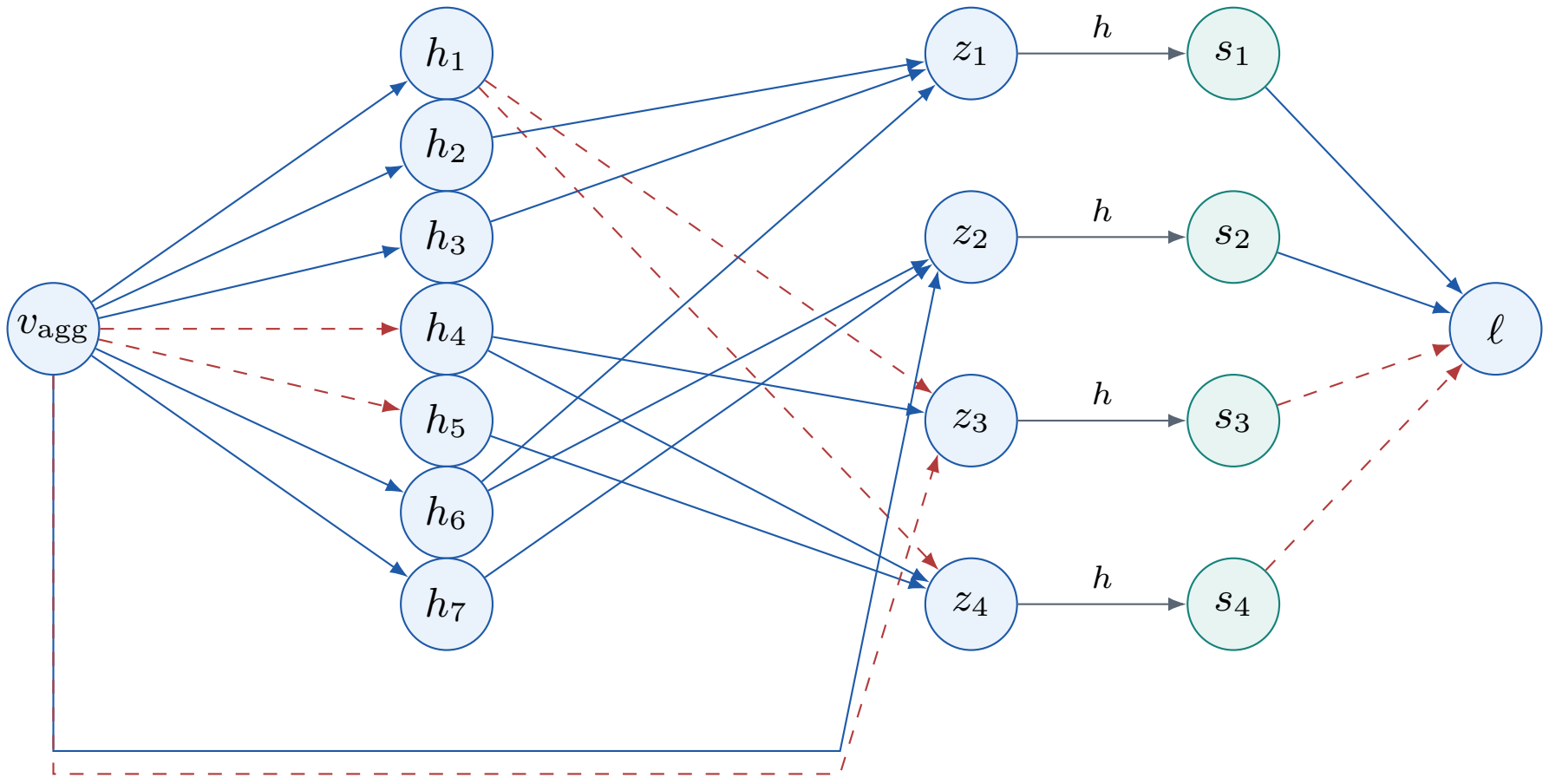
\begin{figure}[htbp]
\centering
\begin{tikzpicture}[x=10mm,y=7mm,font=\small,unit/.style={circle,draw=NPBlue,fill=NPLightBlue,minimum size=7mm,inner sep=1pt},posedge/.style={-{Latex[length=1.4mm]},draw=NPBlue},negedge/.style={-{Latex[length=1.4mm]},draw=NPRed,dashed}]
\node[unit] (input) at (0,3) {$v_{\rm agg}$};
\node[unit] (h1) at (3,6) {$h_1$};
\node[unit] (h2) at (3,5) {$h_2$};
\node[unit] (h3) at (3,4) {$h_3$};
\node[unit] (h4) at (3,3) {$h_4$};
\node[unit] (h5) at (3,2) {$h_5$};
\node[unit] (h6) at (3,1) {$h_6$};
\node[unit] (h7) at (3,0) {$h_7$};
\node[unit] (z1) at (7,6) {$z_1$};
\node[unit,fill=NPLightTeal,draw=NPTeal] (s1) at (9,6) {$s_1$};
\draw[-{Latex[length=1.4mm]},draw=NPGray] (z1)--node[above,font=\scriptsize] {$h$}(s1);
\node[unit] (z2) at (7,4) {$z_2$};
\node[unit,fill=NPLightTeal,draw=NPTeal] (s2) at (9,4) {$s_2$};
\draw[-{Latex[length=1.4mm]},draw=NPGray] (z2)--node[above,font=\scriptsize] {$h$}(s2);
\node[unit] (z3) at (7,2) {$z_3$};
\node[unit,fill=NPLightTeal,draw=NPTeal] (s3) at (9,2) {$s_3$};
\draw[-{Latex[length=1.4mm]},draw=NPGray] (z3)--node[above,font=\scriptsize] {$h$}(s3);
\node[unit] (z4) at (7,0) {$z_4$};
\node[unit,fill=NPLightTeal,draw=NPTeal] (s4) at (9,0) {$s_4$};
\draw[-{Latex[length=1.4mm]},draw=NPGray] (z4)--node[above,font=\scriptsize] {$h$}(s4);
\node[unit] (readout) at (11,3) {$\ell$};
\draw[posedge] (input)--(h1);
\draw[negedge] (h1)--(z3);
\draw[negedge] (h1)--(z4);
\draw[posedge] (input)--(h2);
\draw[posedge] (h2)--(z1);
\draw[posedge] (input)--(h3);
\draw[posedge] (h3)--(z1);
\draw[negedge] (input)--(h4);
\draw[posedge] (h4)--(z3);
\draw[posedge] (h4)--(z4);
\draw[negedge] (input)--(h5);
\draw[posedge] (h5)--(z4);
\draw[posedge] (input)--(h6);
\draw[posedge] (h6)--(z1);
\draw[posedge] (h6)--(z2);
\draw[posedge] (input)--(h7);
\draw[posedge] (h7)--(z2);
\draw[posedge] (input.south)--(0,-1.6)--(6,-1.6)--(z2.south west);
\draw[negedge] (input.south)--(0,-1.85)--(6,-1.85)--(z3.south west);
\draw[posedge] (s1)--(readout);
\draw[posedge] (s2)--(readout);
\draw[negedge] (s3)--(readout);
\draw[negedge] (s4)--(readout);
\end{tikzpicture}
\caption{Measured output cone of the saved circuit. Four identical aggregation inputs are merged into $v_{\rm agg}$ and established constants are folded into biases. Every nonzero hidden-to-output connection is shown; $s_i$ follows thresholding of $z_i$ and $\ell$ is the readout. Solid blue and dashed red edges denote positive and negative coefficients. Table~\ref{tab:measured-weights} supplies the numerical coefficients. The computation affecting the output at two circuit ticks is reduced exactly from the 55-edge saved graph.}
\label{fig:measured-circuit}
\end{figure}

%% file: notation_en.tex
\section{Notation}
\label{app:notation}
The notation below separates circuit topology, input structures, numerical
enclosures, and logical queries. Bound object variables and component indices
are local to their equations. Numeric keys in archived records retain their
original names; the table describes manuscript notation.
\small
\begin{longtable}{@{}p{42mm}p{110mm}@{}}
\toprule
Symbol & Meaning and scope\\
\midrule
\endfirsthead
\toprule
Symbol & Meaning and scope (continued)\\
\midrule
\endhead
\bottomrule
\endfoot
$\mathcal I,\mathcal H,\mathcal O$ & Input, hidden, and output node sets of a local circuit; $|\mathcal H|=12$.\\
$u,v$ & Local source and destination node indices in the circuit equations.\\
$g_{uv},g_v$ & Binary edge and hidden-node gates; $z_e,p_e$ are a gate parameter and its sigmoid probability.\\
$w_{uv},\widehat w_{uv},b_v$ & Saved weight, effective gated weight, and node bias.\\
$a_v^\tau,r_v^\tau$ & Node value and preactivation at circuit tick $\tau$.\\
$F_\theta,K$ & Local numerical function and its circuit-tick budget; $K=2$ in the experiment.\\
$A_{UV},D_{\mathcal H}$ & Effective weight block and diagonal hidden-gate matrix; $I_{\mathrm{out}}$ is the output identity matrix.\\
$v_{\mathrm{aff}},v_{\mathrm{hid}}$ & Affine and hidden vectors in the two-tick reduction.\\
$n_{\mathrm b},n_{\max},f_r$ & Batch structure count, maximum object count in a batch, and feature count at arity $r=0,1,2$.\\
$c_{\mathrm s},S_i^t,\mathsf Q_{ij}^t$ & State width ($c_{\mathrm s}=4$), object states, and pair states.\\
$t,T,\tau$ & Message-tick index, message-tick budget, and local circuit-tick index.\\
$h,h_\alpha$ & State threshold with nonnegative inputs mapped to $+1$, and its training interpolation.\\
$q_{ij},E_i,U_i,N_i$ & Numerical aggregation contribution, existence, universal, and capped-sum operators.\\
$\kappa,n_{+}$ & Numerical sum cap ($\kappa=8$) and number of positive contributions.\\
$J_{ij}^{cd}$ & Pair-state composition through an intermediate object; $c,d$ are component indices.\\
$A,U_A,\Gclass$ & Input structure, its object domain, and the declared class of admissible structures.\\
$r_{\mathrm{tar}},D_{r_{\mathrm{tar}}}(A)$ & Target arity (one or two) and evaluated argument domain.\\
$\Circuit_\theta^{\mathbb R},\Circuit_\theta^{\mathrm{fp}}$ & Real-arithmetic and rounded implementations of the saved program, including outer relational computation.\\
$\psi_\theta,\varphi_A$ & Extracted logical formula and known target rule.\\
$\mathcal Q_1,\mathcal Q_2,\Sigma$ & Logical queries compared by containment and their input constraints.\\
$I=[l,u],\mathcal B,\chi$ & Interval with local lower/upper endpoints, component box, and associated logical condition.\\
$n_{\mathrm v},n_{\mathrm u},\operatorname{Bell}(n_{\mathrm v})$ & Query-variable count, unary-predicate count, and number of variable partitions.\\
$\deg(x),k_{\max},\mathcal D$ & Degree in one relation/direction, largest logical saturation cap, and allowed count set.\\
$F,M,P,S,\operatorname{Bro},V$ & Father, mother, parent, sister, brother, and uncle predicates in the family tasks.\\
$\operatorname{Reach}_k$ & Reachability between distinct vertices by paths of length one through $k$.\\
$\rho(x)$ & Adjacent-to-red formula used as a local abbreviation in the formula examples.\\
$n_{\mathrm{str}},\ell_{\mathrm{lab}},n_{\mathrm{exact}},e_{\mathrm{err}}$ & Structures, labels per structure, completely correct structures, and erroneous labels in Appendix~\ref{app:prediction-detail}.\\
$v_{\rm agg},\eta_k,b_k$ & Shared aggregation input, hidden input coefficient, and folded bias in Appendix~\ref{app:worked-example}.\\
$\delta,\epsilon,W$ & Direct-input coefficient vector, folded output-bias vector, and hidden-to-output matrix in the measured trace.\\
$\omega,\ell,\beta$ & Measured readout weight vector, logit, and bias, with $\ell=\omega^{\mathsf T}s+\beta$.\\
\end{longtable}
\normalsize

%% file: generated/acceptance_stages_en.tex
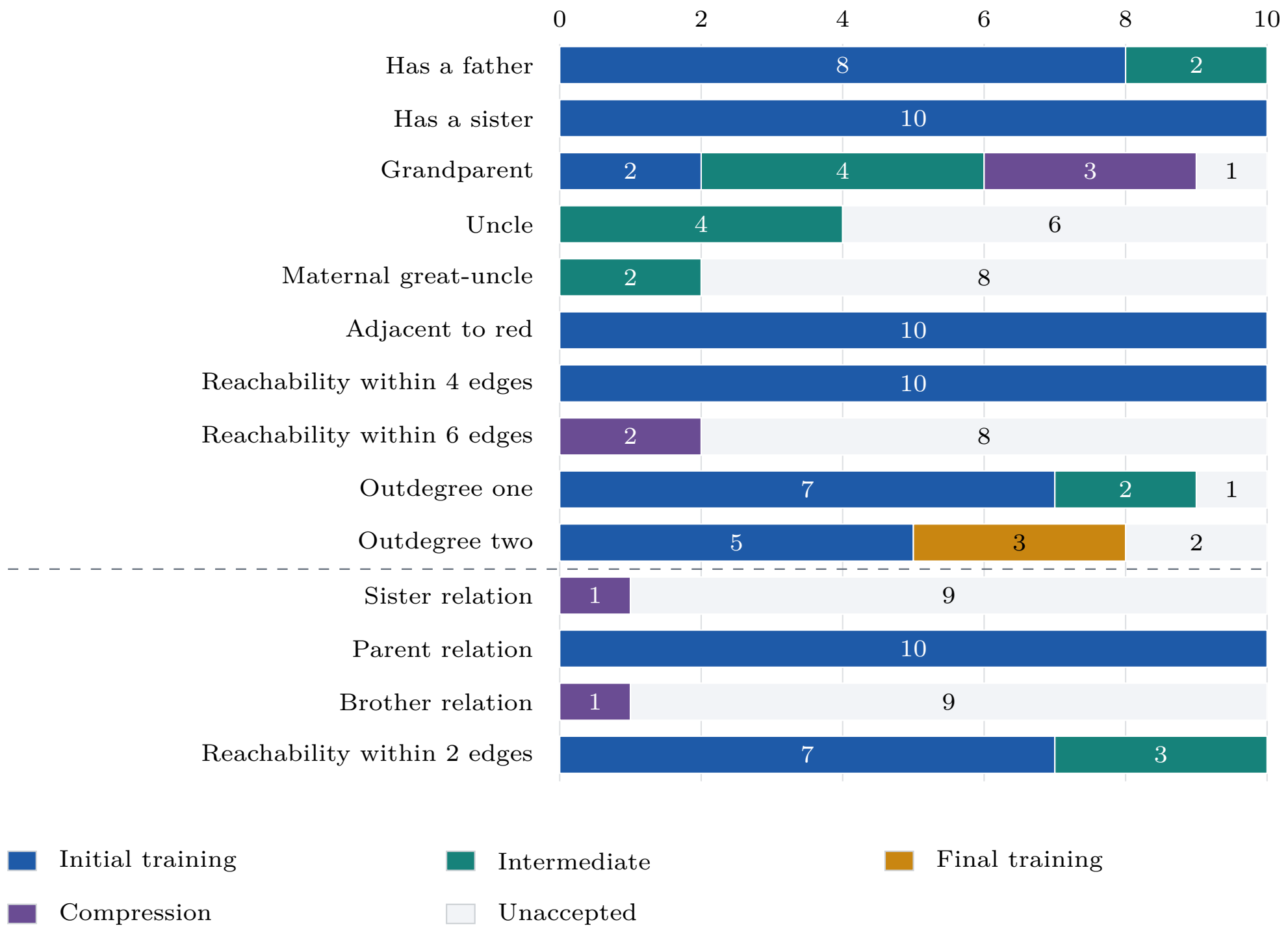
\begin{figure}[htbp]
\centering
\begin{tikzpicture}[x=7.2mm,y=5.4mm,font=\scriptsize]
\draw[NPGray!20] (0,0.4)--(0,-13.5);
\node[above] at (0,0.5) {0};
\draw[NPGray!20] (2,0.4)--(2,-13.5);
\node[above] at (2,0.5) {2};
\draw[NPGray!20] (4,0.4)--(4,-13.5);
\node[above] at (4,0.5) {4};
\draw[NPGray!20] (6,0.4)--(6,-13.5);
\node[above] at (6,0.5) {6};
\draw[NPGray!20] (8,0.4)--(8,-13.5);
\node[above] at (8,0.5) {8};
\draw[NPGray!20] (10,0.4)--(10,-13.5);
\node[above] at (10,0.5) {10};
\node[anchor=east,align=right,text width=57mm] at (-0.2,0) {Has a father};
\filldraw[fill=NPBlue,draw=white] (0,-0.35) rectangle (8,0.35);
\node[text=white] at (4.0,0) {8};
\filldraw[fill=NPTeal,draw=white] (8,-0.35) rectangle (10,0.35);
\node[text=white] at (9.0,0) {2};
\node[anchor=east,align=right,text width=57mm] at (-0.2,-1) {Has a sister};
\filldraw[fill=NPBlue,draw=white] (0,-1.35) rectangle (10,-0.65);
\node[text=white] at (5.0,-1) {10};
\node[anchor=east,align=right,text width=57mm] at (-0.2,-2) {Grandparent};
\filldraw[fill=NPBlue,draw=white] (0,-2.35) rectangle (2,-1.65);
\node[text=white] at (1.0,-2) {2};
\filldraw[fill=NPTeal,draw=white] (2,-2.35) rectangle (6,-1.65);
\node[text=white] at (4.0,-2) {4};
\filldraw[fill=NPPurple,draw=white] (6,-2.35) rectangle (9,-1.65);
\node[text=white] at (7.5,-2) {3};
\filldraw[fill=NPLightGray,draw=white] (9,-2.35) rectangle (10,-1.65);
\node[text=black] at (9.5,-2) {1};
\node[anchor=east,align=right,text width=57mm] at (-0.2,-3) {Uncle};
\filldraw[fill=NPTeal,draw=white] (0,-3.35) rectangle (4,-2.65);
\node[text=white] at (2.0,-3) {4};
\filldraw[fill=NPLightGray,draw=white] (4,-3.35) rectangle (10,-2.65);
\node[text=black] at (7.0,-3) {6};
\node[anchor=east,align=right,text width=57mm] at (-0.2,-4) {Maternal great-uncle};
\filldraw[fill=NPTeal,draw=white] (0,-4.35) rectangle (2,-3.65);
\node[text=white] at (1.0,-4) {2};
\filldraw[fill=NPLightGray,draw=white] (2,-4.35) rectangle (10,-3.65);
\node[text=black] at (6.0,-4) {8};
\node[anchor=east,align=right,text width=57mm] at (-0.2,-5) {Adjacent to red};
\filldraw[fill=NPBlue,draw=white] (0,-5.35) rectangle (10,-4.65);
\node[text=white] at (5.0,-5) {10};
\node[anchor=east,align=right,text width=57mm] at (-0.2,-6) {Reachability within 4 edges};
\filldraw[fill=NPBlue,draw=white] (0,-6.35) rectangle (10,-5.65);
\node[text=white] at (5.0,-6) {10};
\node[anchor=east,align=right,text width=57mm] at (-0.2,-7) {Reachability within 6 edges};
\filldraw[fill=NPPurple,draw=white] (0,-7.35) rectangle (2,-6.65);
\node[text=white] at (1.0,-7) {2};
\filldraw[fill=NPLightGray,draw=white] (2,-7.35) rectangle (10,-6.65);
\node[text=black] at (6.0,-7) {8};
\node[anchor=east,align=right,text width=57mm] at (-0.2,-8) {Outdegree one};
\filldraw[fill=NPBlue,draw=white] (0,-8.35) rectangle (7,-7.65);
\node[text=white] at (3.5,-8) {7};
\filldraw[fill=NPTeal,draw=white] (7,-8.35) rectangle (9,-7.65);
\node[text=white] at (8.0,-8) {2};
\filldraw[fill=NPLightGray,draw=white] (9,-8.35) rectangle (10,-7.65);
\node[text=black] at (9.5,-8) {1};
\node[anchor=east,align=right,text width=57mm] at (-0.2,-9) {Outdegree two};
\filldraw[fill=NPBlue,draw=white] (0,-9.35) rectangle (5,-8.65);
\node[text=white] at (2.5,-9) {5};
\filldraw[fill=NPGold,draw=white] (5,-9.35) rectangle (8,-8.65);
\node[text=black] at (6.5,-9) {3};
\filldraw[fill=NPLightGray,draw=white] (8,-9.35) rectangle (10,-8.65);
\node[text=black] at (9.0,-9) {2};
\node[anchor=east,align=right,text width=57mm] at (-0.2,-10) {Sister relation};
\filldraw[fill=NPPurple,draw=white] (0,-10.35) rectangle (1,-9.65);
\node[text=white] at (0.5,-10) {1};
\filldraw[fill=NPLightGray,draw=white] (1,-10.35) rectangle (10,-9.65);
\node[text=black] at (5.5,-10) {9};
\node[anchor=east,align=right,text width=57mm] at (-0.2,-11) {Parent relation};
\filldraw[fill=NPBlue,draw=white] (0,-11.35) rectangle (10,-10.65);
\node[text=white] at (5.0,-11) {10};
\node[anchor=east,align=right,text width=57mm] at (-0.2,-12) {Brother relation};
\filldraw[fill=NPPurple,draw=white] (0,-12.35) rectangle (1,-11.65);
\node[text=white] at (0.5,-12) {1};
\filldraw[fill=NPLightGray,draw=white] (1,-12.35) rectangle (10,-11.65);
\node[text=black] at (5.5,-12) {9};
\node[anchor=east,align=right,text width=57mm] at (-0.2,-13) {Reachability within 2 edges};
\filldraw[fill=NPBlue,draw=white] (0,-13.35) rectangle (7,-12.65);
\node[text=white] at (3.5,-13) {7};
\filldraw[fill=NPTeal,draw=white] (7,-13.35) rectangle (10,-12.65);
\node[text=white] at (8.5,-13) {3};
\draw[NPGray,dashed] (-7.8,-9.5)--(10,-9.5);
\filldraw[draw=NPGray!30,fill=NPBlue] (-7.8,-15.18) rectangle (-7.3999999999999995,-14.82);
\node[anchor=west] at (-7.25,-15.0) {Initial training};
\filldraw[draw=NPGray!30,fill=NPTeal] (-1.5999999999999996,-15.18) rectangle (-1.1999999999999997,-14.82);
\node[anchor=west] at (-1.0499999999999996,-15.0) {Intermediate};
\filldraw[draw=NPGray!30,fill=NPGold] (4.6000000000000005,-15.18) rectangle (5.000000000000001,-14.82);
\node[anchor=west] at (5.15,-15.0) {Final training};
\filldraw[draw=NPGray!30,fill=NPPurple] (-7.8,-16.18) rectangle (-7.3999999999999995,-15.82);
\node[anchor=west] at (-7.25,-16.0) {Compression};
\filldraw[draw=NPGray!30,fill=NPLightGray] (-1.5999999999999996,-16.18) rectangle (-1.1999999999999997,-15.82);
\node[anchor=west] at (-1.0499999999999996,-16.0) {Unaccepted};
\end{tikzpicture}
\caption{Selection stage of accepted checkpoints. Each bar contains ten trials; the bottom four tasks are auxiliary. Intermediate checkpoints are assessed in priority order after continuation training. Unaccepted combines counterexample and inconclusive outcomes.}
\label{fig:acceptance-stages}
\end{figure}

%% file: generated/compact_formulas_en.tex
\begin{table}[tbp]
\centering\small
\caption{Formula reduction and standard outputs for the selected circuits. Closed class-vocabulary expressions are compared before and after reduction; shorter bodies using named ADF calls become the standard output. All counts include the off-diagonal output guard.}
\label{tab:compact-formulas}
\begin{tabular}{lrrr}
\toprule
Task & Closed & Reduced & Standard \\
\midrule
Has a father & 35 & 23 & 23 \\
Has a sister & 1,249 & 214 & 30$^{*}$ \\
Grandparent & 762 & 158 & 63$^{*}$ \\
Uncle & 267 & 196 & 89$^{*}$ \\
Maternal great-uncle & 320 & 249 & 55$^{*}$ \\
Adjacent to red & 34 & 32 & 32 \\
Reachability within 4 edges & 307 & 307 & 113$^{*}$ \\
Reachability within 6 edges & 747 & 747 & 113$^{*}$ \\
Outdegree one & 42 & 41 & 41 \\
Outdegree two & 42 & 41 & 41 \\
Sister relation & 1,218 & 202 & 202 \\
Parent relation & 284 & 30 & 30 \\
Brother relation & 197 & 126 & 126 \\
Reachability within 2 edges & 80 & 68 & 68 \\
\bottomrule
\end{tabular}
\par\smallskip\parbox{0.98\linewidth}{\footnotesize Columns count characters. UCQ atoms/disjuncts are removed by containment proofs; degree formulas use exact saturated-count profiles. Asterisks mark bodies using named ADFs, with each definition supplied once. Expansion can lengthen a body; size including definitions is recorded separately. Raw formulas, rewrite provenance, and definitions are retained.}
\end{table}

%% file: generated/compact_examples_en.tex
The standard outputs include the following compact programs. The degree expression writes $\min(3,d)=2$ as $d=2$; the reachability expression factors out the common off-diagonal guard.

\begin{align*}
 \rho(x)&=\exists z\,[\operatorname{adj}(x,z)\wedge\operatorname{red}(z)],\\
 \operatorname{Outdegree}_2(x)&=\bigl|\{z:\operatorname{adj}(x,z)\}\bigr|=2,\\
 \widehat{\operatorname{Reach}}_6(x,y)&=(x\ne y)\wedge\bigl[\operatorname{Reach}_4(x,y)\vee
 \exists z\,[\operatorname{Reach}_2(x,z)\wedge\operatorname{Reach}_4(z,y)]\bigr].
\end{align*}
$\operatorname{Reach}_2$ and $\operatorname{Reach}_4$ name acquired ADFs whose definitions accompany the output; binary ADF calls return false on self-pairs. The evidence record reports the body size together with the sizes of its dependency definitions.

%% file: generated/formula_diagnostics_en.tex
\begin{table}[tbp]
\centering\small
\caption{Measured syntax sizes of selected verified formulas. Extracted formulas retain names of ADFs; closed formulas expand them and apply the recorded Boolean minimization. The library column sums the trial body and referenced dependency bodies once, before expansion.}
\label{tab:formula-diagnostics}
\begin{tabular}{lrrrrrr}
\toprule
Task (seed) & Extracted & Closed & Tree & DAG & Depth & Library \\
 & chars. & chars. & nodes & nodes & & chars. \\
\midrule
Adjacent to red (8) & 34 & 34 & 3 & 3 & 2 & 34 \\
Adjacent to red (3) & 355,428 & 34 & 31,766 & 287 & 9 & 355,428 \\
Reachability within 6 edges (0) & 136 & 14,809 & 8 & 6 & 4 & 76,251 \\
\bottomrule
\end{tabular}
\par\smallskip\parbox{0.98\linewidth}{\footnotesize Tree nodes count atoms and logical operators, with n-ary conjunctions/disjunctions counted once; depth counts nodes from root to leaf. DAG nodes share identical ordered subexpressions without further rewriting. Library sizes omit definition headers and implicit off-diagonal guards; the closed formulas include the guards. These metrics quantify formula syntax and description size.}
\end{table}

%% file: generated/formula_lengths_en.tex
\begin{figure}[htbp]
\centering
\begin{tikzpicture}[x=12.50000mm,y=10mm,font=\small]
\draw[NPGray!20] (0,-.4)--(0,2.5);
\node[below] at (0,-.5) {$10^0$};
\draw[NPGray!20] (1,-.4)--(1,2.5);
\node[below] at (1,-.5) {$10^1$};
\draw[NPGray!20] (2,-.4)--(2,2.5);
\node[below] at (2,-.5) {$10^2$};
\draw[NPGray!20] (3,-.4)--(3,2.5);
\node[below] at (3,-.5) {$10^3$};
\draw[NPGray!20] (4,-.4)--(4,2.5);
\node[below] at (4,-.5) {$10^4$};
\draw[NPGray!20] (5,-.4)--(5,2.5);
\node[below] at (5,-.5) {$10^5$};
\draw[NPGray!20] (6,-.4)--(6,2.5);
\node[below] at (6,-.5) {$10^6$};
\draw[NPGray!20] (7,-.4)--(7,2.5);
\node[below] at (7,-.5) {$10^7$};
\draw[NPGray!20] (8,-.4)--(8,2.5);
\node[below] at (8,-.5) {$10^8$};
\node[anchor=east] at (-.15,2) {Verified (96)};
\fill[NPTeal,opacity=.7] (3.30428,1.80000) circle[radius=1.2pt];
\fill[NPTeal,opacity=.7] (1.54407,1.95556) circle[radius=1.2pt];
\fill[NPTeal,opacity=.7] (4.62963,2.11111) circle[radius=1.2pt];
\fill[NPTeal,opacity=.7] (4.62963,1.84444) circle[radius=1.2pt];
\fill[NPTeal,opacity=.7] (4.62963,2.00000) circle[radius=1.2pt];
\fill[NPTeal,opacity=.7] (4.62963,2.15556) circle[radius=1.2pt];
\fill[NPTeal,opacity=.7] (1.54407,1.88889) circle[radius=1.2pt];
\fill[NPTeal,opacity=.7] (4.62963,2.04444) circle[radius=1.2pt];
\fill[NPTeal,opacity=.7] (4.62963,2.20000) circle[radius=1.2pt];
\fill[NPTeal,opacity=.7] (1.54407,1.93333) circle[radius=1.2pt];
\fill[NPTeal,opacity=.7] (1.62325,2.08889) circle[radius=1.2pt];
\fill[NPTeal,opacity=.7] (1.62325,1.82222) circle[radius=1.2pt];
\fill[NPTeal,opacity=.7] (2.14301,1.97778) circle[radius=1.2pt];
\fill[NPTeal,opacity=.7] (1.62325,2.13333) circle[radius=1.2pt];
\fill[NPTeal,opacity=.7] (1.62325,1.86667) circle[radius=1.2pt];
\fill[NPTeal,opacity=.7] (1.62325,2.02222) circle[radius=1.2pt];
\fill[NPTeal,opacity=.7] (1.62325,2.17778) circle[radius=1.2pt];
\fill[NPTeal,opacity=.7] (1.62325,1.91111) circle[radius=1.2pt];
\fill[NPTeal,opacity=.7] (2.14301,2.06667) circle[radius=1.2pt];
\fill[NPTeal,opacity=.7] (1.62325,1.80000) circle[radius=1.2pt];
\fill[NPTeal,opacity=.7] (1.79934,1.95556) circle[radius=1.2pt];
\fill[NPTeal,opacity=.7] (2.63548,2.11111) circle[radius=1.2pt];
\fill[NPTeal,opacity=.7] (3.28058,1.84444) circle[radius=1.2pt];
\fill[NPTeal,opacity=.7] (1.79934,2.00000) circle[radius=1.2pt];
\fill[NPTeal,opacity=.7] (1.79934,2.15556) circle[radius=1.2pt];
\fill[NPTeal,opacity=.7] (1.79934,1.88889) circle[radius=1.2pt];
\fill[NPTeal,opacity=.7] (2.26951,2.04444) circle[radius=1.2pt];
\fill[NPTeal,opacity=.7] (1.79934,2.20000) circle[radius=1.2pt];
\fill[NPTeal,opacity=.7] (1.79934,1.93333) circle[radius=1.2pt];
\fill[NPTeal,opacity=.7] (2.32222,2.08889) circle[radius=1.2pt];
\fill[NPTeal,opacity=.7] (1.86923,1.82222) circle[radius=1.2pt];
\fill[NPTeal,opacity=.7] (1.86923,1.97778) circle[radius=1.2pt];
\fill[NPTeal,opacity=.7] (1.86923,2.13333) circle[radius=1.2pt];
\fill[NPTeal,opacity=.7] (1.74036,1.86667) circle[radius=1.2pt];
\fill[NPTeal,opacity=.7] (2.29447,2.02222) circle[radius=1.2pt];
\fill[NPTeal,opacity=.7] (3.42813,2.17778) circle[radius=1.2pt];
\fill[NPTeal,opacity=.7] (2.38739,1.91111) circle[radius=1.2pt];
\fill[NPTeal,opacity=.7] (3.34380,2.06667) circle[radius=1.2pt];
\fill[NPTeal,opacity=.7] (5.55075,1.80000) circle[radius=1.2pt];
\fill[NPTeal,opacity=.7] (2.01703,1.95556) circle[radius=1.2pt];
\fill[NPTeal,opacity=.7] (5.24968,2.11111) circle[radius=1.2pt];
\fill[NPTeal,opacity=.7] (3.59151,1.84444) circle[radius=1.2pt];
\fill[NPTeal,opacity=.7] (5.24968,2.00000) circle[radius=1.2pt];
\fill[NPTeal,opacity=.7] (1.53148,2.15556) circle[radius=1.2pt];
\fill[NPTeal,opacity=.7] (4.27054,1.88889) circle[radius=1.2pt];
\fill[NPTeal,opacity=.7] (3.94522,2.04444) circle[radius=1.2pt];
\fill[NPTeal,opacity=.7] (3.91677,2.20000) circle[radius=1.2pt];
\fill[NPTeal,opacity=.7] (3.46687,1.93333) circle[radius=1.2pt];
\fill[NPTeal,opacity=.7] (3.78547,2.08889) circle[radius=1.2pt];
\fill[NPTeal,opacity=.7] (4.33620,1.82222) circle[radius=1.2pt];
\fill[NPTeal,opacity=.7] (2.13354,1.97778) circle[radius=1.2pt];
\fill[NPTeal,opacity=.7] (3.13130,2.13333) circle[radius=1.2pt];
\fill[NPTeal,opacity=.7] (3.28803,1.86667) circle[radius=1.2pt];
\fill[NPTeal,opacity=.7] (3.38112,2.02222) circle[radius=1.2pt];
\fill[NPTeal,opacity=.7] (3.81697,2.17778) circle[radius=1.2pt];
\fill[NPTeal,opacity=.7] (2.13354,1.91111) circle[radius=1.2pt];
\fill[NPTeal,opacity=.7] (2.41830,2.06667) circle[radius=1.2pt];
\fill[NPTeal,opacity=.7] (1.62325,1.80000) circle[radius=1.2pt];
\fill[NPTeal,opacity=.7] (1.62325,1.95556) circle[radius=1.2pt];
\fill[NPTeal,opacity=.7] (1.62325,2.11111) circle[radius=1.2pt];
\fill[NPTeal,opacity=.7] (3.01620,1.84444) circle[radius=1.2pt];
\fill[NPTeal,opacity=.7] (3.01620,2.00000) circle[radius=1.2pt];
\fill[NPTeal,opacity=.7] (1.62325,2.15556) circle[radius=1.2pt];
\fill[NPTeal,opacity=.7] (1.62325,1.88889) circle[radius=1.2pt];
\fill[NPTeal,opacity=.7] (1.62325,2.04444) circle[radius=1.2pt];
\fill[NPTeal,opacity=.7] (1.62325,2.20000) circle[radius=1.2pt];
\fill[NPTeal,opacity=.7] (3.28375,1.93333) circle[radius=1.2pt];
\fill[NPTeal,opacity=.7] (1.62325,2.08889) circle[radius=1.2pt];
\fill[NPTeal,opacity=.7] (1.62325,1.82222) circle[radius=1.2pt];
\fill[NPTeal,opacity=.7] (1.62325,1.97778) circle[radius=1.2pt];
\fill[NPTeal,opacity=.7] (2.63949,2.13333) circle[radius=1.2pt];
\fill[NPTeal,opacity=.7] (1.62325,1.86667) circle[radius=1.2pt];
\fill[NPTeal,opacity=.7] (1.62325,2.02222) circle[radius=1.2pt];
\fill[NPTeal,opacity=.7] (1.62325,2.17778) circle[radius=1.2pt];
\fill[NPTeal,opacity=.7] (3.20439,1.91111) circle[radius=1.2pt];
\fill[NPTeal,opacity=.7] (3.76185,2.06667) circle[radius=1.2pt];
\fill[NPTeal,opacity=.7] (4.24497,1.80000) circle[radius=1.2pt];
\fill[NPTeal,opacity=.7] (3.51654,1.95556) circle[radius=1.2pt];
\fill[NPTeal,opacity=.7] (3.73783,2.11111) circle[radius=1.2pt];
\fill[NPTeal,opacity=.7] (3.35717,1.84444) circle[radius=1.2pt];
\fill[NPTeal,opacity=.7] (3.10003,2.00000) circle[radius=1.2pt];
\fill[NPTeal,opacity=.7] (2.92942,2.15556) circle[radius=1.2pt];
\fill[NPTeal,opacity=.7] (4.13027,1.88889) circle[radius=1.2pt];
\fill[NPTeal,opacity=.7] (3.77510,2.04444) circle[radius=1.2pt];
\fill[NPTeal,opacity=.7] (3.36248,2.20000) circle[radius=1.2pt];
\fill[NPTeal,opacity=.7] (2.36549,1.93333) circle[radius=1.2pt];
\fill[NPTeal,opacity=.7] (4.82802,2.08889) circle[radius=1.2pt];
\fill[NPTeal,opacity=.7] (6.67511,1.82222) circle[radius=1.2pt];
\fill[NPTeal,opacity=.7] (2.42160,1.97778) circle[radius=1.2pt];
\fill[NPTeal,opacity=.7] (7.33115,2.13333) circle[radius=1.2pt];
\fill[NPTeal,opacity=.7] (6.05261,1.86667) circle[radius=1.2pt];
\fill[NPTeal,opacity=.7] (2.68034,2.02222) circle[radius=1.2pt];
\fill[NPTeal,opacity=.7] (3.85564,2.17778) circle[radius=1.2pt];
\fill[NPTeal,opacity=.7] (3.93676,1.91111) circle[radius=1.2pt];
\fill[NPTeal,opacity=.7] (6.02081,2.06667) circle[radius=1.2pt];
\fill[NPTeal,opacity=.7] (3.94562,1.80000) circle[radius=1.2pt];
\draw[black,thick] (2.63749,1.7)--(2.63749,2.3);
\node[anchor=east] at (-.15,1) {Counterexample (18)};
\fill[NPRed,opacity=.7] (2.25042,0.80000) circle[radius=1.2pt];
\fill[NPRed,opacity=.7] (0.69897,0.95556) circle[radius=1.2pt];
\fill[NPRed,opacity=.7] (0.69897,1.11111) circle[radius=1.2pt];
\fill[NPRed,opacity=.7] (0.69897,0.84444) circle[radius=1.2pt];
\fill[NPRed,opacity=.7] (0.69897,1.00000) circle[radius=1.2pt];
\fill[NPRed,opacity=.7] (2.87216,1.15556) circle[radius=1.2pt];
\fill[NPRed,opacity=.7] (0.60206,0.88889) circle[radius=1.2pt];
\fill[NPRed,opacity=.7] (2.35793,1.04444) circle[radius=1.2pt];
\fill[NPRed,opacity=.7] (2.56585,1.20000) circle[radius=1.2pt];
\fill[NPRed,opacity=.7] (1.36173,0.93333) circle[radius=1.2pt];
\fill[NPRed,opacity=.7] (2.59218,1.08889) circle[radius=1.2pt];
\fill[NPRed,opacity=.7] (0.60206,0.82222) circle[radius=1.2pt];
\fill[NPRed,opacity=.7] (0.60206,0.97778) circle[radius=1.2pt];
\fill[NPRed,opacity=.7] (0.69897,1.13333) circle[radius=1.2pt];
\fill[NPRed,opacity=.7] (0.60206,0.86667) circle[radius=1.2pt];
\fill[NPRed,opacity=.7] (0.69897,1.02222) circle[radius=1.2pt];
\fill[NPRed,opacity=.7] (2.31175,1.17778) circle[radius=1.2pt];
\fill[NPRed,opacity=.7] (0.69897,0.91111) circle[radius=1.2pt];
\draw[black,thick] (0.69897,0.7)--(0.69897,1.3);
\node[anchor=east] at (-.15,0) {Inconclusive (26)};
\fill[NPGold,opacity=.7] (2.01703,-0.20000) circle[radius=1.2pt];
\fill[NPGold,opacity=.7] (3.03503,-0.04444) circle[radius=1.2pt];
\fill[NPGold,opacity=.7] (2.59660,0.11111) circle[radius=1.2pt];
\fill[NPGold,opacity=.7] (2.41996,-0.15556) circle[radius=1.2pt];
\fill[NPGold,opacity=.7] (2.42813,0.00000) circle[radius=1.2pt];
\fill[NPGold,opacity=.7] (2.19312,0.15556) circle[radius=1.2pt];
\fill[NPGold,opacity=.7] (2.09342,-0.11111) circle[radius=1.2pt];
\fill[NPGold,opacity=.7] (3.69311,0.04444) circle[radius=1.2pt];
\fill[NPGold,opacity=.7] (1.82607,0.20000) circle[radius=1.2pt];
\fill[NPGold,opacity=.7] (2.34830,-0.06667) circle[radius=1.2pt];
\fill[NPGold,opacity=.7] (1.86923,0.08889) circle[radius=1.2pt];
\fill[NPGold,opacity=.7] (2.73799,-0.17778) circle[radius=1.2pt];
\fill[NPGold,opacity=.7] (1.86923,-0.02222) circle[radius=1.2pt];
\fill[NPGold,opacity=.7] (2.77670,0.13333) circle[radius=1.2pt];
\fill[NPGold,opacity=.7] (3.53958,-0.13333) circle[radius=1.2pt];
\fill[NPGold,opacity=.7] (2.69020,0.02222) circle[radius=1.2pt];
\fill[NPGold,opacity=.7] (6.06890,0.17778) circle[radius=1.2pt];
\fill[NPGold,opacity=.7] (3.07188,-0.08889) circle[radius=1.2pt];
\fill[NPGold,opacity=.7] (2.58206,0.06667) circle[radius=1.2pt];
\fill[NPGold,opacity=.7] (4.34741,-0.20000) circle[radius=1.2pt];
\fill[NPGold,opacity=.7] (3.94660,-0.04444) circle[radius=1.2pt];
\fill[NPGold,opacity=.7] (3.05231,0.11111) circle[radius=1.2pt];
\fill[NPGold,opacity=.7] (2.87622,-0.15556) circle[radius=1.2pt];
\fill[NPGold,opacity=.7] (3.61300,0.00000) circle[radius=1.2pt];
\fill[NPGold,opacity=.7] (4.33586,0.15556) circle[radius=1.2pt];
\fill[NPGold,opacity=.7] (3.22814,-0.11111) circle[radius=1.2pt];
\draw[black,thick] (2.75778,-0.3)--(2.75778,0.3);
\node[below] at (4.0,-1.05) {Characters in the extracted formula};
\end{tikzpicture}
\caption{Extracted-formula lengths for the selected circuits in all 140 trials. The logarithmic axis counts saved characters before expanding ADFs or further simplification. Each point is one trial; black ticks mark group medians and parentheses give sample sizes. Vertical offsets only reduce overplotting.}
\label{fig:formula-lengths}
\end{figure}
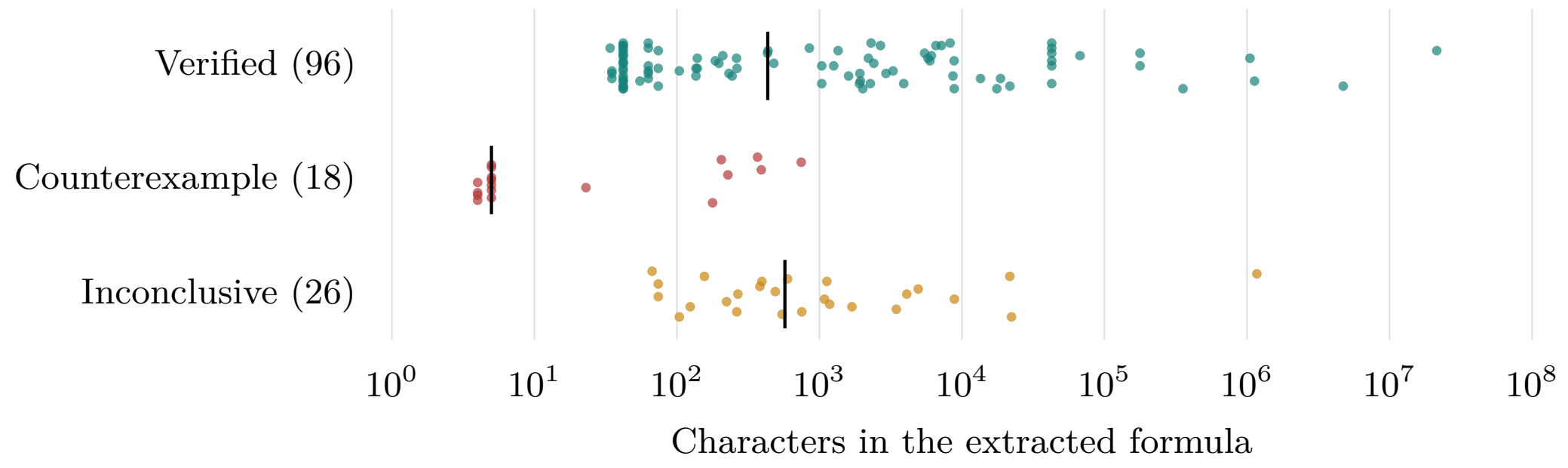

%% file: generated/detector_protocol_en.tex
\FloatBarrier
\section{Detector Pilot Protocol and Complete Local Checking}
\label{app:detector-protocol}
\subsection{Tasks, matched conditions, and optimization}
The bounded neural experiment in Section~\ref{sec:detector-experiment} uses directed loop-free graphs with two binary targets:
\[
 \varphi_{\mathrm{rev}}(x,y)=\operatorname{adj}(y,x),\qquad
 \varphi_{\mathrm{sym}}(x,y)=\operatorname{adj}(x,y)\vee\operatorname{adj}(y,x).
\]
Outputs are evaluated at $x\ne y$. Seeds 0--9 are matched across four detectors and feedback disabled or enabled, giving $2\times10\times4\times2=160$ training runs. For a given task and seed, all eight conditions share initialization and initial supervised data. Neural NP4G has one state channel, three hidden nodes, two circuit ticks, one message tick, and no relational-composition primitive. Its inputs retain adjacency, identity, and the unary \texttt{red} feature, so candidate rules can depend on information irrelevant to the target.

The initial dataset is one three-object graph. Each directed off-diagonal edge is independently present with probability $0.25$, and each object is red with probability $0.5$. Such small data leave several rules consistent with the observations, exposing the contribution of detector information. Adam uses learning rate $0.02$, a gradient-norm limit of $1.0$, and at most 600 updates. The hard-state coefficient increases linearly to one during the first 100 updates. There is no pruning. Every 200 updates, the detector checks a candidate in the same optimizer trajectory; acceptance stops the run. A detector query is therefore a checkpoint attempt, whereas a seed is an independent initialization.

The objective retains class-balanced cross entropy on the initial data, adds one feedback loss when available, and adds $10^{-4}\mathcal L_{\mathrm{structure}}$, where the structure term is the model's edge/node regularizer. Positive and negative labels receive equal total weight when both occur. Feedback examples accumulate in a buffer and are selected cyclically, one per subsequent optimizer update, with loss coefficient one. Labeled feedback supervises only the single returned output cell with binary cross entropy. Let $p_A(x,y)$ be the sigmoid of the output logit on graph $A$. A symmetry counterexample contributes
\[
 \bigl(p_A(x,y)-p_A(y,x)\bigr)^2,
\]
and a monotonicity counterexample with $A\subseteq B$ and unchanged unary features contributes
\[
 \max\{0,p_A(x,y)-p_B(x,y)\}^2.
\]
Property feedback supplies no target label. Feedback-disabled runs continue on the initial data. A counterexample returned at the last budget boundary contributes no later update and is not counted as replayed.

\subsection{Detector information and acceptance}
The formula detector holds and exposes the target formula; the black-box detector retains it internally and returns only the decision and one labeled counterexample. Both perform exact equivalence checking after faithful extraction. Their common optimization policy uses training examples and returned counterexamples, without consulting target syntax. They therefore control access to the formula while preserving the oracle's semantic information. Their paired trajectories are identical.

The partial detector specifies edge monotonicity for both tasks and output symmetry for symmetrization. Boolean binary typing is fixed by the model. It returns a violated property instance without a target label and accepts when all specified properties hold under the complete local check below. Both constant-false and constant-true rules satisfy these properties, demonstrating that this specification is strictly weaker than either target.

The finite detector uses a target evaluator on eight four-object graphs, independently generated for its fixed panel. For each graph, an edge probability is drawn uniformly from $[0.05,0.5]$, directed edges are sampled independently, the diagonal is cleared, and unary red labels have probability $0.5$. It returns the first labeled disagreement and accepts when none is found. Thus partial and finite settings vary different aspects of detection: the former supplies weaker properties with complete checking in the local fragment; the latter supplies target labels with a finite search budget.

\subsection{Completeness of local checking}
After faithful extraction, a candidate in the pilot's quantifier-free local fragment depends only on
\[
 \operatorname{adj}(x,y),\quad\operatorname{adj}(y,x),\quad
 \operatorname{red}(x),\quad\operatorname{red}(y).
\]
Equality is false at evaluated pairs. Each of the $2^4=16$ assignments is realized on two distinct objects. Conversely, every evaluated pair in any larger loop-free graph induces one of these assignments, and a local formula cannot inspect the remaining objects. Equality of the candidate and target on all 16 assignments therefore establishes their equivalence at every evaluated pair for arbitrary object counts. The exact service uses the existing containment checker or this complete truth table; the local fallback applies only to the stated quantifier-free grammar.

The partial checker tests symmetry on the same assignments. To check edge monotonicity, each of the two directed edge bits has three inclusion-consistent states $(0,0),(0,1),(1,1)$ across a graph and its supergraph. Holding the two unary bits fixed gives $3^2\times2^2=36$ comparisons, covering every local edge-inclusion case. Unsupported extraction or grammar yields an inconclusive result, preserving the stated acceptance conditions.

\subsection{Audit, query counts, and evidence}
An independent exact audit follows the stopping decision and is withheld from training. Detector acceptance, exact equivalence of the final candidate, and acquisition are recorded separately; acquisition requires both acceptance and equivalence. The audit reloaded all 160 saved models, repeated extraction and equivalence checking, and compared saved-circuit and formula outputs on all 16 local assignments, evaluating both ordered distinct pairs. All 5,120 output cells agreed. Records preserve the formulas, counterexamples, checkpoints, and source and checkpoint hashes.

Table~\ref{tab:detector-study} averages query counts across all ten runs, including budget exhaustion and erroneous acceptance. Under exact detection, feedback changes these means from 2.0 to 1.9 for reversal and from 2.6 to 2.4 for symmetrization. Among successful runs alone, means increase from 1.0 to 1.625 and from 1.0 to 1.8, respectively: feedback adds successes at later checkpoints. For partial specifications, feedback changes accepted counts from 8/10 to 10/10 on both tasks, while correct acquisitions change from 5/10 to 6/10 on reversal and remain 2/10 on symmetrization. The finite detector's 7/10 correct symmetrization acquisitions with feedback exceed the exact detector's 5/10 under the same update budget; their returned counterexamples differ.

%% file: references.bib
@article{angluin1987learning,
  author = {Angluin, Dana},
  title = {Learning Regular Sets from Queries and Counterexamples},
  journal = {Information and Computation},
  volume = {75}, number = {2}, pages = {87--106}, year = {1987},
  doi = {10.1016/0890-5401(87)90052-6}
}

@inproceedings{solarlezama2006sketching,
  author = {Solar-Lezama, Armando and Tancau, Liviu and Bodik, Rastislav and Saraswat, Vijay and Seshia, Sanjit},
  title = {Combinatorial Sketching for Finite Programs},
  booktitle = {Proceedings of the 12th International Conference on Architectural Support for Programming Languages and Operating Systems},
  pages = {404--415}, year = {2006},
  doi = {10.1145/1168857.1168907}
}

@article{gehring1993error,
  author = {Gehring, William J. and Goss, Brian and Coles, Michael G. H. and Meyer, David E. and Donchin, Emanuel},
  title = {A Neural System for Error Detection and Compensation},
  journal = {Psychological Science}, volume = {4}, number = {6},
  pages = {385--390}, year = {1993},
  doi = {10.1111/j.1467-9280.1993.tb00586.x}
}

@inproceedings{hara2024np4g,
  author    = {Hara, Shoichiro and Watanabe, Yuji},
  title     = {{NP4G}: Network Programming for Generalization},
  editor    = {Arai, Kohei},
  booktitle = {Intelligent Systems and Applications},
  series    = {Lecture Notes in Networks and Systems},
  volume    = {822},
  pages     = {301--315},
  publisher = {Springer},
  year      = {2024},
  doi       = {10.1007/978-3-031-47721-8_20}
}

@inproceedings{dong2019nlm,
  author    = {Dong, Honghua and Mao, Jiayuan and Lin, Tian and Wang, Chong and Li, Lihong and Zhou, Denny},
  title     = {Neural Logic Machines},
  booktitle = {International Conference on Learning Representations},
  year      = {2019},
  url       = {https://openreview.net/forum?id=B1xY-hRctX}
}

@article{zimmer2023dlm,
  author  = {Zimmer, Matthieu and Feng, Xuening and Glanois, Claire and Jiang, Zhaohui and Zhang, Jianyi and Weng, Paul and Li, Dong and Hao, Jianye and Liu, Wulong},
  title   = {Differentiable Logic Machines},
  journal = {Transactions on Machine Learning Research},
  year    = {2023},
  url     = {https://openreview.net/forum?id=mXfkKtu5JA}
}

@article{evans2018dilp,
  author  = {Evans, Richard and Grefenstette, Edward},
  title   = {Learning Explanatory Rules from Noisy Data},
  journal = {Journal of Artificial Intelligence Research},
  volume  = {61},
  pages   = {1--64},
  year    = {2018},
  doi     = {10.1613/jair.5714}
}

@inproceedings{yang2017neurallp,
  author    = {Yang, Fan and Yang, Zhilin and Cohen, William W.},
  title     = {Differentiable Learning of Logical Rules for Knowledge Base Reasoning},
  booktitle = {Advances in Neural Information Processing Systems},
  volume    = {30},
  pages     = {2319--2328},
  year      = {2017}
}

@inproceedings{manhaeve2018deepproblog,
  author    = {Manhaeve, Robin and Duman{\v{c}}i{\'c}, Sebastijan and Kimmig, Angelika and Demeester, Thomas and De Raedt, Luc},
  title     = {{DeepProbLog}: Neural Probabilistic Logic Programming},
  booktitle = {Advances in Neural Information Processing Systems},
  volume    = {31},
  pages     = {3753--3763},
  year      = {2018}
}

@inproceedings{rocktaschel2017proving,
  author    = {Rockt{\"a}schel, Tim and Riedel, Sebastian},
  title     = {End-to-End Differentiable Proving},
  booktitle = {Advances in Neural Information Processing Systems},
  volume    = {30},
  pages     = {3788--3800},
  year      = {2017}
}

@inproceedings{petersen2022dlogic,
  author    = {Petersen, Felix and Borgelt, Christian and Kuehne, Hilde and Deussen, Oliver},
  title     = {Deep Differentiable Logic Gate Networks},
  booktitle = {Advances in Neural Information Processing Systems},
  volume    = {35},
  pages     = {2006--2018},
  year      = {2022}
}

@inproceedings{jacovi2020faithfulness,
  author    = {Jacovi, Alon and Goldberg, Yoav},
  title     = {Towards Faithfully Interpretable {NLP} Systems: How Should We Define and Evaluate Faithfulness?},
  booktitle = {Proceedings of the 58th Annual Meeting of the Association for Computational Linguistics},
  pages     = {4198--4205},
  year      = {2020},
  doi       = {10.18653/v1/2020.acl-main.386}
}

@article{lipton2018mythos,
  author  = {Lipton, Zachary C.},
  title   = {The Mythos of Model Interpretability},
  journal = {Communications of the ACM},
  volume  = {61},
  number  = {10},
  pages   = {36--43},
  year    = {2018},
  doi     = {10.1145/3233231}
}

@inproceedings{katz2017reluplex,
  author    = {Katz, Guy and Barrett, Clark and Dill, David L. and Julian, Kyle and Kochenderfer, Mykel J.},
  title     = {{Reluplex}: An Efficient {SMT} Solver for Verifying Deep Neural Networks},
  booktitle = {Computer Aided Verification},
  series    = {Lecture Notes in Computer Science},
  volume    = {10426},
  pages     = {97--117},
  publisher = {Springer},
  year      = {2017},
  doi       = {10.1007/978-3-319-63387-9_5}
}

@article{bengio2013ste,
  author  = {Bengio, Yoshua and L{\'e}onard, Nicholas and Courville, Aaron},
  title   = {Estimating or Propagating Gradients Through Stochastic Neurons for Conditional Computation},
  journal = {arXiv preprint arXiv:1308.3432},
  year    = {2013},
  doi     = {10.48550/arXiv.1308.3432}
}

@article{rockafellar2000cvar,
  author  = {Rockafellar, R. Tyrrell and Uryasev, Stanislav},
  title   = {Optimization of Conditional Value-at-Risk},
  journal = {Journal of Risk},
  volume  = {2},
  number  = {3},
  pages   = {21--41},
  year    = {2000},
  doi     = {10.21314/JOR.2000.038}
}

@inproceedings{chandra1977cq,
  author    = {Chandra, Ashok K. and Merlin, Philip M.},
  title     = {Optimal Implementation of Conjunctive Queries in Relational Data Bases},
  booktitle = {Proceedings of the Ninth Annual ACM Symposium on Theory of Computing},
  pages     = {77--90},
  year      = {1977},
  doi       = {10.1145/800105.803397}
}

@inproceedings{cousot1977abstract,
  author    = {Cousot, Patrick and Cousot, Radhia},
  title     = {Abstract Interpretation: A Unified Lattice Model for Static Analysis of Programs by Construction or Approximation of Fixpoints},
  booktitle = {Proceedings of the Fourth ACM SIGACT-SIGPLAN Symposium on Principles of Programming Languages},
  pages     = {238--252},
  year      = {1977},
  url       = {https://www.di.ens.fr/~cousot/publications.www/CousotCousot-POPL-77-ACM-p238--252-1977.pdf}
}

@manual{mpfrmanual,
  title        = {{GNU MPFR} Manual},
  organization = {The MPFR Team},
  year         = {2026},
  note         = {Accessed September 12, 2026. Rounding modes and correctly rounded arithmetic},
  url          = {https://www.mpfr.org/mpfr-current/mpfr.html}
}

@article{muggleton1991ilp,
  author  = {Muggleton, Stephen},
  title   = {Inductive Logic Programming},
  journal = {New Generation Computing},
  volume  = {8},
  pages   = {295--318},
  year    = {1991},
  doi     = {10.1007/BF03037089}
}

@book{koza1994gp2,
  author    = {Koza, John R.},
  title     = {Genetic Programming {II}: Automatic Discovery of Reusable Programs},
  publisher = {MIT Press},
  year      = {1994},
  isbn      = {9780262111898},
  url       = {https://mitpress.mit.edu/9780262111898/genetic-programming-ii/}
}

@inproceedings{ellis2021dreamcoder,
  author    = {Ellis, Kevin and Wong, Catherine and Nye, Maxwell and Sabl{\'e}-Meyer, Mathias and Morales, Lucas and Hewitt, Luke and Cary, Luc and Solar-Lezama, Armando and Tenenbaum, Joshua B.},
  title     = {{DreamCoder}: Bootstrapping Inductive Program Synthesis with Wake-Sleep Library Learning},
  booktitle = {Proceedings of the 42nd ACM SIGPLAN International Conference on Programming Language Design and Implementation},
  pages     = {835--850},
  publisher = {ACM},
  year      = {2021},
  doi       = {10.1145/3453483.3454080}
}

@article{stanley2002neat,
  author  = {Stanley, Kenneth O. and Miikkulainen, Risto},
  title   = {Evolving Neural Networks through Augmenting Topologies},
  journal = {Evolutionary Computation},
  volume  = {10},
  number  = {2},
  pages   = {99--127},
  year    = {2002},
  doi     = {10.1162/106365602320169811}
}

@inproceedings{wei2022cot,
  author    = {Wei, Jason and Wang, Xuezhi and Schuurmans, Dale and Bosma, Maarten and Ichter, Brian and Xia, Fei and Chi, Ed and Le, Quoc V. and Zhou, Denny},
  title     = {Chain-of-Thought Prompting Elicits Reasoning in Large Language Models},
  booktitle = {Advances in Neural Information Processing Systems},
  volume    = {35},
  pages     = {24824--24837},
  year      = {2022},
  url       = {https://proceedings.neurips.cc/paper/2022/hash/9d5609613524ecf4f15af0f7b31abca4-Abstract-Conference.html}
}

@article{guo2025deepseekr1,
  author  = {Guo, Daya and Yang, Dejian and Zhang, Haowei and others},
  title   = {{DeepSeek-R1} Incentivizes Reasoning in {LLMs} through Reinforcement Learning},
  journal = {Nature},
  volume  = {645},
  pages   = {633--638},
  year    = {2025},
  doi     = {10.1038/s41586-025-09422-z},
  url     = {https://www.nature.com/articles/s41586-025-09422-z}
}

@inproceedings{mirzadeh2025gsm,
  author    = {Mirzadeh, Iman and Alizadeh-Vahid, Keivan and Shahrokhi, Hooman and Tuzel, Oncel and Bengio, Samy and Farajtabar, Mehrdad},
  title     = {{GSM-Symbolic}: Understanding the Limitations of Mathematical Reasoning in Large Language Models},
  booktitle = {International Conference on Learning Representations},
  year      = {2025},
  url       = {https://proceedings.iclr.cc/paper_files/paper/2025/hash/ec2e7a896f8250986b3907f57621ce94-Abstract-Conference.html}
}

@inproceedings{dziri2023faith,
  author    = {Dziri, Nouha and Lu, Ximing and Sclar, Melanie and Li, Xiang (Lorraine) and Jiang, Liwei and Lin, Bill Yuchen and Welleck, Sean and West, Peter and Bhagavatula, Chandra and Le Bras, Ronan and Hwang, Jena and Sanyal, Soumya and Ren, Xiang and Ettinger, Allyson and Harchaoui, Zaid and Choi, Yejin},
  title     = {Faith and Fate: Limits of Transformers on Compositionality},
  booktitle = {Advances in Neural Information Processing Systems},
  volume    = {36},
  pages     = {70293--70332},
  year      = {2023},
  url       = {https://papers.neurips.cc/paper_files/paper/2023/hash/deb3c28192f979302c157cb653c15e90-Abstract-Conference.html}
}

@article{trinh2024alphageometry,
  author  = {Trinh, Trieu H. and Wu, Yuhuai and Le, Quoc V. and He, He and Luong, Thang},
  title   = {Solving Olympiad Geometry without Human Demonstrations},
  journal = {Nature},
  volume  = {625},
  pages   = {476--482},
  year    = {2024},
  doi     = {10.1038/s41586-023-06747-5},
  url     = {https://www.nature.com/articles/s41586-023-06747-5}
}

@inproceedings{yao2026compositional,
  author    = {Yao, Xinhao and Ren, Ruifeng and Liao, Yun and Ding, Lizhong and Liu, Yong},
  title     = {Compositional Generalization from Learned Skills via {CoT} Training: A Theoretical and Structural Analysis for Reasoning},
  booktitle = {International Conference on Learning Representations},
  year      = {2026},
  url       = {https://proceedings.iclr.cc/paper_files/paper/2026/hash/9990a18179e862042c18dfc856b0e53f-Abstract-Conference.html}
}

@inproceedings{barcelo2020logical,
  author = {Barcel{\'o}, Pablo and Kostylev, Egor V. and Monet, Mika{\"e}l and P{\'e}rez, Jorge and Reutter, Juan L. and Silva, Juan Pablo},
  title = {The Logical Expressiveness of Graph Neural Networks},
  booktitle = {International Conference on Learning Representations},
  year = {2020},
  url = {https://ai.dmi.unibas.ch/research/reading_group/barcelo-et-al-iclr2020.pdf}
}

@inproceedings{grohe2021logic,
  author = {Grohe, Martin},
  title = {The Logic of Graph Neural Networks},
  booktitle = {36th Annual ACM/IEEE Symposium on Logic in Computer Science},
  pages = {1--17},
  year = {2021},
  doi = {10.1109/LICS52264.2021.9470677},
  url = {https://arxiv.org/abs/2104.14624}
}

@inproceedings{salzer2023limits,
  author = {S{\"a}lzer, Marco and Lange, Martin},
  title = {Fundamental Limits in Formal Verification of Message-Passing Neural Networks},
  booktitle = {International Conference on Learning Representations},
  year = {2023},
  url = {https://openreview.net/forum?id=WlbG820mRH-}
}

@inproceedings{benedikt2024decidability,
  author = {Benedikt, Michael and Lu, Chia-Hsuan and Motik, Boris and Tan, Tony},
  title = {Decidability of Graph Neural Networks via Logical Characterizations},
  booktitle = {51st International Colloquium on Automata, Languages, and Programming},
  series = {Leibniz International Proceedings in Informatics},
  volume = {297},
  pages = {127:1--127:20},
  year = {2024},
  doi = {10.4230/LIPIcs.ICALP.2024.127},
  url = {https://drops.dagstuhl.de/entities/document/10.4230/LIPIcs.ICALP.2024.127}
}

@article{towell1993rules,
  author = {Towell, Geoffrey G. and Shavlik, Jude W.},
  title = {Extracting Refined Rules from Knowledge-Based Neural Networks},
  journal = {Machine Learning},
  volume = {13},
  pages = {71--101},
  year = {1993},
  doi = {10.1007/BF00993103},
  url = {https://link.springer.com/article/10.1007/BF00993103}
}

@article{andrews1995survey,
  author = {Andrews, Robert and Diederich, Joachim and Tickle, Alan B.},
  title = {Survey and Critique of Techniques for Extracting Rules from Trained Artificial Neural Networks},
  journal = {Knowledge-Based Systems},
  volume = {8},
  number = {6},
  pages = {373--389},
  year = {1995},
  doi = {10.1016/0950-7051(96)81920-4},
  url = {https://www.sciencedirect.com/science/article/abs/pii/0950705196819204}
}

@inproceedings{weiss2018automata,
  author = {Weiss, Gail and Goldberg, Yoav and Yahav, Eran},
  title = {Extracting Automata from Recurrent Neural Networks Using Queries and Counterexamples},
  booktitle = {Proceedings of the 35th International Conference on Machine Learning},
  series = {Proceedings of Machine Learning Research},
  volume = {80},
  pages = {5247--5256},
  year = {2018},
  url = {https://proceedings.mlr.press/v80/weiss18a.html}
}

@article{gowal2018ibp,
  author = {Gowal, Sven and Dvijotham, Krishnamurthy and Stanforth, Robert and Bunel, Rudy and Qin, Chongli and Uesato, Jonathan and Arandjelovi{\'c}, Relja and Mann, Timothy and Kohli, Pushmeet},
  title = {On the Effectiveness of Interval Bound Propagation for Training Verifiably Robust Models},
  journal = {arXiv preprint arXiv:1810.12715},
  year = {2018},
  url = {https://arxiv.org/abs/1810.12715}
}

@inproceedings{mirman2018diffai,
  author = {Mirman, Matthew and Gehr, Timon and Vechev, Martin},
  title = {Differentiable Abstract Interpretation for Provably Robust Neural Networks},
  booktitle = {Proceedings of the 35th International Conference on Machine Learning},
  series = {Proceedings of Machine Learning Research},
  volume = {80},
  pages = {3578--3586},
  year = {2018},
  url = {https://proceedings.mlr.press/v80/mirman18b.html}
}

@inproceedings{han2015connections,
  author = {Han, Song and Pool, Jeff and Tran, John and Dally, William J.},
  title = {Learning Both Weights and Connections for Efficient Neural Network},
  booktitle = {Advances in Neural Information Processing Systems},
  volume = {28},
  pages = {1135--1143},
  year = {2015},
  url = {https://proceedings.neurips.cc/paper/2015/hash/ae0eb3eed39d2bcef4622b2499a05fe6-Abstract.html}
}

@inproceedings{louizos2018lzero,
  author = {Louizos, Christos and Welling, Max and Kingma, Diederik P.},
  title = {Learning Sparse Neural Networks through {$L_0$} Regularization},
  booktitle = {International Conference on Learning Representations},
  year = {2018},
  url = {https://openreview.net/forum?id=H1Y8hhg0b}
}

@inproceedings{bellec2018rewiring,
  author = {Bellec, Guillaume and Kappel, David and Maass, Wolfgang and Legenstein, Robert},
  title = {Deep Rewiring: Training Very Sparse Deep Networks},
  booktitle = {International Conference on Learning Representations},
  year = {2018},
  url = {https://openreview.net/forum?id=BJ_wN01C-}
}

@inproceedings{verma2018pirl,
  author = {Verma, Abhinav and Murali, Vijayaraghavan and Singh, Rishabh and Kohli, Pushmeet and Chaudhuri, Swarat},
  title = {Programmatically Interpretable Reinforcement Learning},
  booktitle = {Proceedings of the 35th International Conference on Machine Learning},
  series = {Proceedings of Machine Learning Research},
  volume = {80},
  pages = {5045--5054},
  year = {2018},
  url = {https://proceedings.mlr.press/v80/verma18a.html}
}

@inproceedings{bastani2018viper,
  author = {Bastani, Osbert and Pu, Yewen and Solar-Lezama, Armando},
  title = {Verifiable Reinforcement Learning via Policy Extraction},
  booktitle = {Advances in Neural Information Processing Systems},
  volume = {31},
  year = {2018},
  url = {https://proceedings.neurips.cc/paper_files/paper/2018/hash/e6d8545daa42d5ced125a4bf747b3688-Abstract.html}
}

@article{cropper2022ilp,
  author = {Cropper, Andrew and Duman{\v c}i{\'c}, Sebastijan},
  title = {Inductive Logic Programming at 30: A New Introduction},
  journal = {Journal of Artificial Intelligence Research},
  volume = {74},
  pages = {765--850},
  year = {2022},
  doi = {10.1613/jair.1.13507},
  url = {https://doi.org/10.1613/jair.1.13507}
}

@article{cropper2021popper,
  author = {Cropper, Andrew and Morel, Rolf},
  title = {Learning Programs by Learning from Failures},
  journal = {Machine Learning},
  volume = {110},
  pages = {801--856},
  year = {2021},
  doi = {10.1007/s10994-020-05934-z},
  url = {https://link.springer.com/article/10.1007/s10994-020-05934-z}
}

@inproceedings{levy1993updates,
  author = {Levy, Alon Y. and Sagiv, Yehoshua},
  title = {Queries Independent of Updates},
  booktitle = {Proceedings of the 19th International Conference on Very Large Data Bases},
  pages = {171--181},
  year = {1993},
  url = {https://www.vldb.org/conf/1993/P171.PDF}
}

@inproceedings{wei2003negation,
  author = {Wei, Fang and Lausen, Georg},
  title = {Containment of Conjunctive Queries with Safe Negation},
  booktitle = {Database Theory---ICDT 2003},
  series = {Lecture Notes in Computer Science},
  volume = {2572},
  pages = {343--357},
  year = {2003},
  doi = {10.1007/3-540-36285-1_23},
  url = {https://link.springer.com/chapter/10.1007/3-540-36285-1_23}
}

@article{clopper1934limits,
  author = {Clopper, C. J. and Pearson, E. S.},
  title = {The Use of Confidence or Fiducial Limits Illustrated in the Case of the Binomial},
  journal = {Biometrika},
  volume = {26},
  number = {4},
  pages = {404--413},
  year = {1934},
  doi = {10.1093/biomet/26.4.404},
  url = {https://doi.org/10.1093/biomet/26.4.404}
}

@article{novikov2025alphaevolve,
  author  = {Novikov, Alexander and V{\~u}, Ng{\^a}n and Eisenberger, Marvin and Dupont, Emilien and Huang, Po-Sen and Wagner, Adam Zsolt and others},
  title   = {{AlphaEvolve}: A Coding Agent for Scientific and Algorithmic Discovery},
  journal = {arXiv preprint arXiv:2506.13131},
  year    = {2025},
  doi     = {10.48550/arXiv.2506.13131},
  url     = {https://arxiv.org/abs/2506.13131}
}

@article{hubert2026alphaproof,
  author  = {Hubert, Thomas and Mehta, Rishi and Sartran, Laurent and others},
  title   = {Olympiad-Level Formal Mathematical Reasoning with Reinforcement Learning},
  journal = {Nature},
  volume  = {651},
  pages   = {607--613},
  year    = {2026},
  doi     = {10.1038/s41586-025-09833-y},
  url     = {https://www.nature.com/articles/s41586-025-09833-y}
}

@techreport{chaudhuri2026alphaproofcode,
  author      = {Chaudhuri, Swarat and Ferraiuolo, Andrew and Kovsharov, Anton and Chajed, Tej and Schlesinger, Sam and Seth, Karn and Katz, Jonathan and Sutton, Charles and Kohli, Pushmeet},
  title       = {{AlphaProof Code}: Verifying {Signal}'s Security at the Machine-Code Level},
  institution = {Google DeepMind and Google},
  year        = {2026},
  month       = nov,
  url         = {https://deepmind.google/documents/36/Formal_Verification_of_Signal_4.pdf},
  note        = {Technical report dated 10 November 2026}
}

@inproceedings{xu2026dafnycomp,
  author    = {Xu, Xu and Li, Xin and Qu, Xingwei and Fu, Jie and Yuan, Binhang},
  title     = {Local Success Does Not Compose: Benchmarking Large Language Models for Compositional Formal Verification},
  booktitle = {International Conference on Learning Representations},
  year      = {2026},
  url       = {https://proceedings.iclr.cc/paper_files/paper/2026/hash/c04d37be05ba74419d2d5705972a9d64-Abstract-Conference.html}
}
